\documentclass[10pt,journal,compsoc]{IEEEtran}

\ifCLASSOPTIONcompsoc
  \usepackage[nocompress]{cite}
\else
  \usepackage{cite}
\fi

\usepackage{microtype}
\usepackage{graphicx}
\usepackage{subfigure}
\usepackage{booktabs} 

\usepackage{amsfonts}
\usepackage{epstopdf}
\usepackage{algorithm}

\usepackage{bm}
\usepackage{amsmath}
\usepackage{graphicx}
\usepackage{slashbox}
\usepackage{ulem}
\usepackage{nicefrac}       
\usepackage{color}

\usepackage{algorithmic}

\usepackage{amsthm}
\newtheorem{theorem}{Theorem}
\newtheorem{corollary}{Corollary}

\usepackage{hyperref}

\begin{document}

\title{Domain Adaptation with Optimal Transport on the Manifold of SPD matrices}

\author{Or~Yair,
        Felix Dietrich,
        Ioannis G. Kevrekidis,
        and~Ronen Talmon
\IEEEcompsocitemizethanks{\IEEEcompsocthanksitem O. Yair and R. Talmon are with the Viterbi Faculty of Electrical Engineering, Technion, Israel Institute of Technology.\protect\\
E-mail: oryair@campus.technion.ac.il
\IEEEcompsocthanksitem Felix Dietrich is with the Technical University of Munich.
\IEEEcompsocthanksitem Ioannis G. Kevrekidis is with the Department of Chemical and Biomolecular Engineering and of Applied Mathematics and Statistics, Johns Hopkins University.}
\thanks{The work of O. Yair and R. Talmon was supported by the European Union's Horizon 2020 research grant agreement 802735. The work of F. Dietrich and I. G. Kevrekidis was partially supported by DARPA and by the Army Research Office MURI.}
}




\IEEEtitleabstractindextext{%
\begin{abstract}
    In this paper, we address the problem of Domain Adaptation (DA) using Optimal Transport (OT) on Riemannian manifolds. We model the difference between two domains by a diffeomorphism and use the polar factorization theorem to claim that OT is indeed optimal for DA in a well-defined sense, up to a volume preserving map. We then focus on the manifold of Symmetric and Positive-Definite (SPD) matrices, whose structure provided a useful context in recent applications. We demonstrate the polar factorization theorem on this manifold. Due to the uniqueness of the weighted Riemannian mean, and by exploiting existing regularized OT algorithms, we formulate a simple algorithm that maps the source domain to the target domain. We test our algorithm on two Brain-Computer Interface (BCI) data sets and observe state of the art performance.
\end{abstract}

}

\maketitle

\IEEEdisplaynontitleabstractindextext
\IEEEpeerreviewmaketitle

\IEEEraisesectionheading{\section{Introduction}\label{sec:introduction}}
\IEEEPARstart{I}{n}  many applications, the acquired data sets do not reside in the same domain. 
Many factors can contribute to this problem; below we outline just a few notable examples that often are encountered in applications involving measured data.
(i) Acquisition systems and equipment: different sets are obtained by similar but not identical sensors, possibly stemming from different calibrations. (ii) Settings and configurations: when measuring physiological signals, environmental parameters such as the room temperature or the time of day affect the recorded signals.
(iii) Different sites and subjects: in the BCI tasks we consider in the present work, different data sets are recorded from different subjects performing similar tasks.  
DA methods are developed for alleviating the differences between the data sets stemming from such factors \cite{daume2006domain,gong2012geodesic,pan2009survey}.

{\color{black}
To make the problem and setting more concrete, consider two sets of measurements. 
Suppose the two sets have similar content, but are significantly different in their representation due to various nuisance factors such as the ones mentioned above. 
The goal of DA is to find a new representation of the two sets, such that any subsequent processing and analysis applied to their \emph{union} is invariant to their differences and is at least as informative as the application to each set separately.}


Here, we focus on the problem of DA on the cone manifold of SPD matrices. This particular interest arises because SPD matrices have proven to be useful features {\color{black}in a broad range of fields such as machine learning, computer vision, pattern recognition, and biomedical imaging};
see for example \cite{pennec2006riemannian,arsigny2007geometric,wang2012covariance,jayasumana2013kernel,barachant2013classification} and references therein. In addition, SPD matrices represent a broad family; while the use of SPD \textit{covariance} matrices is perhaps the most prominent and widely-spread, there exist other important types of SPD matrices for data analysis such as kernel matrices, diffusion tensor images, graph-related operators, and many more \cite{wang2015beyond}.
Naturally, the problem of DA carries over from the data to their (SPD) features \cite{freifeld2014model}.

In this paper, we follow a recent line of work (see for example \cite{courty2017optimal} and references therein) and propose to solve an OT problem in order to obtain a transportation plan which matches between the
{\color{black}
data distributions of the two sets.
Specifically, one set is considered the source set, and we find a transportation plan that maps it to the domain of the other set, which is considered the target set.
}
While the vast majority of methods solve and implement OT in the Euclidean space, in this work we propose to consider OT on the cone manifold of SPD matrices. Due to the properties of the SPD cone manifold, the transportation plan induces a unique mapping between the two sets, which in turn prescribes an easy-to-implement algorithm for DA on the SPD cone manifold.
This method can be carried out in a purely unsupervised manner, {\color{black}where both the source and the target sets are label-free. 
In addition, we also consider the case where samples in the source set are associated with labels. In this case, we incorporate an
appropriate regularization term to the OT problem, which exploits the knowledge of the labels in the source set \cite{courty2014domain}.
}
Importantly, the proposed method can be applied in a similar manner to other manifolds such as the Grassmannian manifold or the Stiefel manifold, which are also useful {\color{black}embedding of features} for high-dimensional data representation.

The analysis of the proposed approach is based on the polar factorization theorem, which provides an informative view on the usage of OT for DA.
Particularly, we show that the polar factorization theorem implies that OT is indeed optimal for DA in a well-defined sense: it can recover the correct transformation up to a volume preserving map. This result highlights the advantages of this approach, but also, and perhaps more importantly, it establishes a fundamental limitation that applies to every DA method in an unsupervised setting that takes into account only the data densities. 
We illustrate the theoretical results on toy examples and show state of the art numerical results on two BCI data sets.

In summary, our main contributions are as follows.
(i) Based on the polar factorization theorem for diffeomorphisms on Riemannian manifolds, we present a new analysis, highlighting the advantages and limitations of OT for DA. Our analysis includes the general Riemannian setting with the particular SPD manifold as a special case.
(ii) We propose a well-defined and easy to implement algorithm for DA with OT on the manifold of SPD matrices. 
(iii) We show that the applications of the proposed algorithm to two BCI data sets achieve state of the art results.

\section{Preliminaries}
\subsection{The Cone Manifold of SPD Matrices}
    \label{sub:RiemannianGeometry}
    We give a brief overview of the Riemannian geometry of the cone manifold of SPD matrices. For more details, we refer the reader to \cite{bhatia2009positive}. A real symmetric matrix $\boldsymbol{P}\in \mathbb{R}^{d\times d}$ is positive-definite if and only if it has only strictly positive eigenvalues. 
    The collection of all SPD matrices constitutes a convex half-cone in the vector space of real $d\!\times\!d$ symmetric matrices. This cone forms a differentiable Riemannian manifold ${\mathcal{P}}_d$ equipped with the following inner product at the tangent space $\mathcal{T}_{\boldsymbol{P}}{\mathcal{P}}_d$ (the space of symmetric matrices) at the point $\boldsymbol{P}\!\in\!{\mathcal{P}}_d$
    \begin{equation}\label{eq:inner_product}
    \big\langle \boldsymbol{A},\boldsymbol{B}\big\rangle _{\mathcal{T}_{\boldsymbol{P}}{\mathcal{P}}_d}=\big\langle \boldsymbol{P}^{-\frac{1}{2}}\boldsymbol{A}\boldsymbol{P}^{-\frac{1}{2}},\boldsymbol{P}^{-\frac{1}{2}}\boldsymbol{B}\boldsymbol{P}^{-\frac{1}{2}}\big\rangle, 
    \end{equation}
    where $\boldsymbol{A},\boldsymbol{B}\in\mathcal{T}_{\boldsymbol{P}}\mathcal{P}_d$ are symmetric matrices, and $\langle \cdot,\cdot\rangle $ is the Euclidean inner product operation.

    $\mathcal{P}_d$ is a Hadamard manifold, namely, it is simply connected and a complete Riemannian manifold of non-positive sectional curvature. Manifolds with non-positive curvature have a unique geodesic between any two points. The geodesic between $\boldsymbol{P} \in \mathcal{P}_d$, and $\boldsymbol{Q}\in\mathcal{P}_d$ is given by
    \begin{align}
    \label{eq:Geodesic}
    \varphi(t) & = \boldsymbol{P}^{\frac{1}{2}}\big(\boldsymbol{P}^{-\frac{1}{2}}\boldsymbol{Q}\boldsymbol{P}^{-\frac{1}{2}}\big)^{t}\boldsymbol{P}^{\frac{1}{2}},\qquad0\leq t\leq1,
    \end{align}
    see \cite{bhatia2009positive}. The arc-length of the geodesic curve defines the following Riemannian distance
    
    \begin{align}
    \label{eq:RiemannianMetric}
    \begin{split}
    d_{R}^{2}\left(\boldsymbol{P,}\boldsymbol{Q}\right) & =\left\Vert \log\left(\boldsymbol{Q}^{-\frac{1}{2}}\boldsymbol{P}\boldsymbol{Q}^{-\frac{1}{2}}\right)\right\Vert _{F}^{2}, \boldsymbol{P},\boldsymbol{Q}\in\mathcal{P}_d
    \end{split}
    \end{align}

    where, $\Vert \cdot\Vert _{F}$ is the Frobenius norm, and $\log(\boldsymbol{P})$ is the matrix logarithm.
    For more details on $\mathcal{P}_d$, please see Appendix A.

\subsection{Optimal Transport on Riemannian Manifolds}
    \label{sub:OptimalTransport}
Let $(\mathcal{M},g)$ be a smooth, connected, oriented, $d$-dimensional Riemannian manifold with metric $g$.
Let $c(p,q)$ be the ``cost of moving a unit of mass'' from point $p$ to point $q$ on ${\mathcal{M}}$, and define two finite Borel measures $\mu_1,\mu_2$ that are absolutely continuous with respect to the volume form of ${\mathcal{M}}$ so that they have densities $f_1,f_2$. Then, the Kantorovich OT problem consists of finding a transport plan $\gamma^*:{\mathcal{M}}\times {\mathcal{M}}\to \mathbb{R}$ that solves
\begin{equation}\label{eq:K optimal transport problem}
\inf_\gamma \int_{{\mathcal{M}}\times {\mathcal{M}}} c(p,q)d\gamma(p,q).
\end{equation}
The infimum ranges over all plans $\gamma$ with $f_2(p)=\int_{\mathcal{M}} \gamma(p,\cdot)d\text{vol}$ and $f_1(q)=\int_{\mathcal{M}} \gamma(\cdot,q)d\text{vol}$. In our case, we choose the cost function $c(p,q):=d_R^2(p,q)$, where $d_R(p,q)$ is the Riemannian distance between two points $p,q\in {\mathcal{M}}$ induced by $g$. This cost function is well-studied in the theory of OT, see \cite{villani-2009,fathi-2010}. In particular, the given assumptions on the measures $\mu_1,\mu_2$ result in a unique solution $\gamma^*$ to \eqref{eq:K optimal transport problem} that is concentrated on the graph of an invertible function $t:{\mathcal{M}}\to {\mathcal{M}}$ such that $\mu_2(V)=\mu_1( t^{-1}(V))$ for all measurable sets $V\subset\mathcal{M}$.

If the densities of the two measures are sampled at $N_1$ and $N_2$ discrete points respectively, they can be represented by vectors $\hat{\boldsymbol{f}}_1\in\mathbb{R}^{N_1},\hat{\boldsymbol{f}}_2\in\mathbb{R}^{N_2}$. This leads to the discrete version of \eqref{eq:K optimal transport problem}:
\begin{equation}\label{eq:discrete ot}
\min_{\boldsymbol{\Gamma} \in \mathcal{F}}\left\langle \boldsymbol{\Gamma}, \boldsymbol{C}\right\rangle,
\end{equation} 
where $\mathcal{F}=\left\{ \boldsymbol{\Gamma}\in\mathbb{R}^{N_{1}\times N_{2}}\,\bigg|\,\boldsymbol{\Gamma}\boldsymbol{1}_{N_{2}}=\hat{\boldsymbol{f}}_{1},\,\boldsymbol{\Gamma}^{T}\boldsymbol{1}_{N_{1}}=\hat{\boldsymbol{f}}_{2}\right\} $.
$\boldsymbol{C}\in\mathbb{R}^{N_{1}\times N_{2}}$
represents the transport cost between the $N_1$ points in the source set and the $N_2$ points in the target set.

\section{Optimal Transport for Domain Adaptation}

\subsection{Proposed Algorithm}
    \label{sub:ProposedMethod}

{\color{black}
Let  $\mathcal{P}=\left\{ \boldsymbol{P}_{i}\in \mathcal{M}\right\} _{i=1}^{N_{1}}$ be a source set, and let $\mathcal{Q}=\left\{ \boldsymbol{Q}_{j}\in \mathcal{M}\right\} _{j=1}^{N_{2}}$ be a target set.
The two sets lie in the same space $\mathcal{M}$ but are concentrated in different parts of it, for example, as a result of sampling from different probability density functions with different supports in $\mathcal{M}$. Consequently, we consider the two sets to be in different domains.



Following \cite{courty2017optimal}, for the purpose of DA, we propose to use OT in order to map the source set $\mathcal{P}$ to the target set $\mathcal{Q}$.
The proposed algorithm appears in Algorithm \ref{alg:OT}.
Let ${\boldsymbol{p}}\in\mathbb{R}^{N_{1}}$ be such that ${\boldsymbol{p}}\left[i\right]=\frac{1}{N_{1}}$ for all $i\in\left\{ 1,2,\dots,N_{1}\right\} $, and similarly let ${\boldsymbol{q}}\in\mathbb{R}^{N_{2}}$ be such that ${\boldsymbol{q}}\left[j\right]=\frac{1}{N_{2}}$ for all $j\in\left\{ 1,2,\dots,N_{2}\right\} $. Set the entries of the OT cost matrix $\boldsymbol{C}\in\mathbb{R}^{N_{1}\times N_{2}}$ as $\boldsymbol{C}\left[i,j\right]=d_{R}^{2}\left(\boldsymbol{P}_{i},\boldsymbol{Q}_{j}\right)$.

Computing $\boldsymbol{p}$ and $\boldsymbol{q}$ using uniform mass distribution is common practice. Yet, for real data sets of empirical observations, using Kernel Density Estimation (KDE) methods may assign weights smaller than $\nicefrac{1}{N}$ to outliers far away from the true target samples. These smaller weights conveniently attenuate the unwanted transport effect caused by these erroneous samples, leading to a more robust algorithm. Concretely, Step 1 may be replaced by  $\boldsymbol{p}\left[i\right]=\frac{1}{Z}\sum_{j=1}^{N_1}\exp\left(-\frac{d_{R}^{2}\left(\boldsymbol{P}_{i},\boldsymbol{P}_{j}\right)}{2\sigma^{2}}\right)$, where $Z\in\mathbb{R}$ is set such that $\sum_{i=1}^{N_{1}}\boldsymbol{p}\left[i\right]=1$ and $\sigma^2$ is usually set around the median of $\left\{ d_{R}^{2}\left(\boldsymbol{P}_{i},\boldsymbol{P}_{j}\right)\right\} _{i,j}$.
In Step 2, $\boldsymbol{q}$ is modified in an analogous fashion.

With the above $\boldsymbol{p},\boldsymbol{q}$ and $\boldsymbol{C}$, we solve (\ref{eq:discrete ot}) and obtain the transport plan $\boldsymbol{\Gamma}$.
For efficient implementation, we propose to obtain the transport plan $\boldsymbol{\Gamma}$ by using the Sinkhorn OT algorithm presented by \cite{cuturi2013sinkhorn}, which extends \eqref{eq:discrete ot} and solves:
\begin{equation}
    \label{eq:Sinkhorn}
    \min_{\boldsymbol{\Gamma}\in \mathcal{F}}\left\langle \boldsymbol{\Gamma},\boldsymbol{C}\right\rangle -\frac{1}{\lambda}h\left(\boldsymbol{\Gamma}\right)    
\end{equation}
where $h\left(\boldsymbol{\Gamma}\right)=-\sum_{i=1}^{N_{1}}\sum_{j=1}^{N_{2}}\boldsymbol{\Gamma}\left[i,j\right]\log\left(\boldsymbol{\Gamma}\left[i,j\right]\right)$ is the entropy of $\boldsymbol{\Gamma}$.
See Appendix B for implementation details.

Step 4 results in the transport plan matrix $\boldsymbol{\Gamma}\in\mathbb{R}^{N_{1}\times N_{2}}$.
From $\boldsymbol{\Gamma}$, we derive the barycentric mapping, $\gamma:\mathcal{M}\to\mathcal{M}$, which serves as our transport map.
Specifically, in Step 5 we compute
\begin{equation}
    \label{eq:barycentric}
    \widetilde{\boldsymbol{P}}_{i}=\gamma\left(\boldsymbol{P}_{i}\right)=\underset{\boldsymbol{P}\in\mathcal{P}_{d}}{\arg\min}\sideset{}{_{j=1}^{N_{2}}}\sum\boldsymbol{\Gamma}\left[i,j\right]d_{R}^{2}\left(\boldsymbol{P},\boldsymbol{Q}_{j}\right).
\end{equation}
In the Euclidean space, (\ref{eq:barycentric}) coincides with the mapping proposed in \cite{courty2017optimal} and is well-defined.
However, when considering a general Riemannian manifold, this quantity might not be well-defined, since the Riemannian mean is not necessarily unique.
Fortunately, in $\mathcal{P}_d$, the Riemannian weighted mean \textit{is} unique and the optimization problem is strictly convex, see \cite{bhatia2009positive}. Therefore, Step 5 of Algorithm \ref{alg:OT} results in a well-defined map.

The weighted mean problem (\ref{eq:barycentric}) can be solved using gradient based algorithms. See Appendix A for the implementation details.
In practice, since $\boldsymbol{\Gamma}$ tends to be sparse, in order to reduce the computational load in Step 5 of Algorithm \ref{alg:OT}, one can use only the highest values in the $i$th row of $\boldsymbol{\Gamma}$.

The map $\gamma$ has two properties which make it useful for DA. (i)
Since $\gamma$ is derived from a solution of OT, $\widetilde{\mathcal{P}}=\gamma\left(\mathcal{P}\right)$ and $\mathcal{Q}$ have similar supports.
(ii) The mapping $\widetilde{\mathcal{P}}=\gamma\left(\mathcal{P}\right)$ maintains the topology of $\mathcal{P}$, i.e., nearby samples in $\mathcal{P}$ remain nearby in $\widetilde{\mathcal{P}}$, because the OT cost is set as squared distance and therefore splitting nearby samples is penalized.

In the case where the source set is associated with labels, we supplement the objective function of the OT \eqref{eq:Sinkhorn} with a regularization term proposed by \cite{courty2014domain} that encourages $\gamma$ to map source samples with the same label closer together.
In this case, Step 4 can be replaced by solving:
\begin{equation}
    \label{eq:RegOT}
    \min_{\boldsymbol{\Gamma} \in \mathcal{F}}\left\langle \boldsymbol{\Gamma},\boldsymbol{C}\right\rangle -\frac{1}{\lambda}h\left(\boldsymbol{\Gamma}\right)+\eta\sum_{j=1}^{N_{2}}\sum_{y=1}^{\left|\mathcal{Y}\right|}\left\Vert \boldsymbol{\Gamma}\left(\mathcal{I}_{y},j\right)\right\Vert _{q}^{p}
\end{equation}
where $\mathcal{Y}$ is the set of all possible labels, $\mathcal{I}_{y}$ is a set containing the indices of the source samples associated with the label $y\in\mathcal{Y}$, $\boldsymbol{\Gamma}\left(\mathcal{I}_{y},j\right)$ is a vector consisting of entries from the $j$th column of $\boldsymbol{\Gamma}$ associated with the label $y$ (i.e., from rows $\mathcal{I}_{y}$) and $\left\Vert \cdot\right\Vert _{q}^{p}$ is the $L_q$ norm to the power of $p$. See Appendix B for more details.

}

\begin{algorithm}
    \renewcommand{\thealgorithm}{1}
    \caption{DA using OT for SPD matrices}
    \label{alg:OT}

    \textbf{\uline{Input}:} source and target sets $\left\{ \boldsymbol{P}_{i}\right\} _{i=1}^{N_{1}}$, $\left\{ \boldsymbol{Q}_{j}\right\} _{j=1}^{N_{2}}$ in $\mathcal{P}_d$.
        
    \textbf{\uline{Output}}\textbf{:} the adapted source set $\left\{ \widetilde{\boldsymbol{P}}_{i}\right\} _{i=1}^{N_{1}}$.
    
    \begin{algorithmic}[1]
        
        \STATE \textbf{for all} $i$ \textbf{do}: $\boldsymbol{p}\left[i\right] \gets \nicefrac{1}{N_{1}}$.
        \hfill \COMMENT{$i\in\left\{ 1,\dots,N_{1}\right\}$}
        
        \STATE \textbf{for all} $j$ \textbf{do}: $\boldsymbol{q}\left[j\right] \gets \nicefrac{1}{N_{2}}$.
        \hfill \COMMENT{$j\in\left\{ 1,\dots,N_{2}\right\}$}
        
        \STATE \textbf{for all} $i$ and $j$ \textbf{do}: $\boldsymbol{C}\left[i,j\right] \gets d_{R}^{2}\left(\boldsymbol{P}_{i},\boldsymbol{Q}_{j}\right)$.
        \hfill \COMMENT{Eq. \eqref{eq:RiemannianMetric}}
        
        \STATE \textbf{set}: $\boldsymbol{\Gamma} \gets \text{SinkhornOptimalTransport}\left(\boldsymbol{p},\boldsymbol{q},\boldsymbol{C}\right)$. \\
        \hfill \COMMENT{see Section \ref{sub:ProposedMethod}}
        
        \STATE \textbf{for all} $i$ \textbf{do}: 
        \hfill \COMMENT{see Section \ref{sub:ProposedMethod}}
        $$\widetilde{\boldsymbol{P}}_{i}\gets\gamma\left(\boldsymbol{P}_{i}\right)=\arg\min_{\boldsymbol{P}\in\mathcal{P}_{d}}\sum_{j=1}^{N_{2}}\boldsymbol{\Gamma}\left[i,j\right]d_{R}^{2}\left(\boldsymbol{P},\boldsymbol{Q}_{j}\right)$$

    \end{algorithmic}
\end{algorithm} 

\subsection{Analysis of DA in Continuous Setting using Polar Factorization: 
Definition and Limitation}
    \label{sub:polarFactorization}

In a continuous setting, following the description in Section \ref{sub:OptimalTransport} leading to \eqref{eq:K optimal transport problem}, we replace the two sets $\mathcal{P}$ and $\mathcal{Q}$ with the two measures $\mu_1$ and $\mu_2$.
We assume that the measure $\mu_2$ is related to $\mu_1$ through a diffeomorphism $s:\mathcal{M}\to \mathcal{M}$ such that for all measurable subsets $V\subset\mathcal{M}$, we have
\begin{equation}\label{eq:mu12 relation} 
\mu_2(V)=\mu_1(s^{-1}(V)).
\end{equation}
In the context of DA, the map $s$ represents the discrepancy between the source domain and the target domain, which we try to bridge. With these prerequisites, we ask:
\textit{given only $\mu_1$ and $\mu_2$ (or their densities $f_1$ and $f_2$, respectively), how much of the information in $s$ can be recovered?}

McCann \cite{mccann2001polar} proved a theorem stating that diffeomorphisms $s$ on Riemannian manifolds have a polar factorization
\begin{equation}\label{eq:mccann decomposition}
s=t\circ u,
\end{equation}
where $t$ is the solution to an OT problem between the two measures $\mu_1$ and $\mu_2$ that are related by \eqref{eq:mu12 relation}, and $u:{\mathcal{M}}\to {\mathcal{M}}$ is a volume-preserving map such that $\mu_1=\mu_1\circ u^{-1}$. The decomposition~\eqref{eq:mccann decomposition} is unique, because the OT problem has a unique solution $t$, and thus $u=t^{-1}\circ s$ is also unique (uniqueness is defined up to sets of measure zero, see \cite{mccann2001polar} for compact manifolds and \cite{fathi-2010} for noncompact manifolds).
It can further be shown that $t$  is the solution to OT if and only if it can be expressed as the gradient of a function $\psi$ that is $c$-convex with respect to the cost function $c$ (see~\cite{brenier-1991,mccann2001polar,modin-2017}).

\begin{figure}
    \centering
    \includegraphics[width=1\columnwidth]{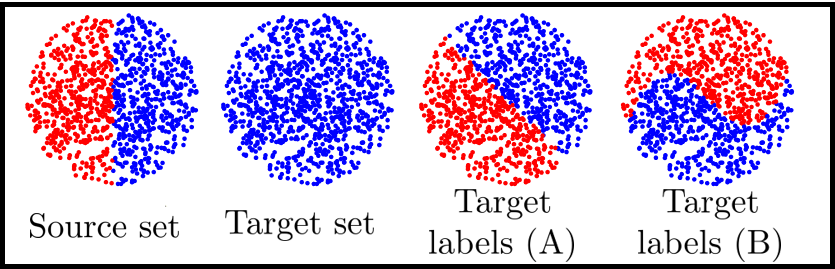}
    \caption{{\color{black}Consider the above (labeled) source set and (unlabeled) target set. Due to rotational symmetries, recovering the true map $s$ (leading to either target labels A or B) by applying OT based only on the target density is impossible. McCann's polar factorization theorem implies that this limitation is fundamental and applies in general to DA in label-free settings}.}
    \label{fig:DA}
\end{figure}

The polar factorization \eqref{eq:mccann decomposition} enables us to present a new analysis for DA using OT.
Broadly, McCann's theorem implies that when we only have access to the densities, the map $s = t \circ u$, which represents the difference between the source and target domains, can be recovered up to a volume preserving function by employing OT. This fundamental limitation is \textit{common to all methods relying on empirical densities} for label-free domain adaptation as illustrated in Figure \ref{fig:DA}. 
More specifically, if for a given map $s$, the map $u$ in \eqref{eq:mccann decomposition} is the identity, solving the OT problem recovers the map $s$. However, if $u$ is not the identity map, solving the OT problem will only recover the function $t$. This means that the correct target density will be recovered, but the approach may still fail to provide the correct point-wise mapping $s$ between the source and the target domains, and hence fail to provide the correct labels on the target domain. The further the map $u$ is away from the identity map, the more distortion is introduced into the adaptation. If the map $u$ is known, such a ``distance'' from the identity map can be measured, for example, in the $L^2$-norm on the space of smooth functions $C^\infty(\mathcal{M})$ on the manifold.

\subsection{Domain Adaptation in $\mathbb{R}^d$}
    \label{sub:EuclideanCase}
When $\mathcal{M}=\mathbb{R}^d$ equipped with the standard Euclidean 2-norm, and if the map $s$ is linear, McCann's polar factorization theorem reduces to polar decomposition of matrices (see~\cite{brenier-1991}), namely:
\begin{equation}\label{eq:polar_dec}
    s\left(\boldsymbol{x}\right)=\boldsymbol{S}\boldsymbol{x}=\boldsymbol{T}\boldsymbol{U}\boldsymbol{x}
\end{equation}
where $\boldsymbol{x}\in\mathbb{R}^{d}$, $\boldsymbol{S}\in\mathbb{R}^{d\times d}$  is invertible, $\boldsymbol{T}=\left(\boldsymbol{S}\boldsymbol{S}^{T}\right)^{\frac{1}{2}} > 0$, and $\boldsymbol{U}=\boldsymbol{T}^{-1}\boldsymbol{S}$ is orthogonal.

An analogous result to the celebrated result of Brenier on the unit sphere \cite{brenier1987decomposition,brenier-1991} was proven in ~\cite{courty2017optimal}:
\begin{theorem}[Discrete case \cite{courty2017optimal}]
\label{thm:discrete}
    Let $\mu_1$ and $\mu_2$ be two discrete distributions with $N$ Diracs.
    If the following conditions hold. 
    (i) The source sample in $\boldsymbol{x}_{i}\in\mathbb{R}^{d},\ \forall i\in\left\{ 1,2,\dots N\right\} $ such that $\boldsymbol{x}_{i}\neq\boldsymbol{x}_{j}$ if $i\neq j$.
    (ii) All weights in the source and target distributions are $\frac{1}{N}$.
    (iii) The target samples are defined as $\boldsymbol{z}_{i}=\boldsymbol{T}\boldsymbol{x}+\boldsymbol{b}$, where $\boldsymbol{T}>0$
    (iv) The cost function is $c\left(\boldsymbol{x},\boldsymbol{z}\right)=\left\Vert \boldsymbol{x}-\boldsymbol{z}\right\Vert _{2}^{2}$
    then, the solution $\gamma$ of the OT problem satisfies $\gamma\left(\boldsymbol{x}_{i}\right)=\boldsymbol{T}\boldsymbol{x}_{i}+\boldsymbol{b}=\boldsymbol{z}_{i},\forall i\in\left\{ 1,2,\dots,N\right\} $.
\end{theorem}
In short, the map $t\left(\boldsymbol{x}\right)=\boldsymbol{T}\boldsymbol{x}+\boldsymbol{b}$ (where $\boldsymbol{T} > 0$) is the solution to an OT problem between an atomic measure $\mu_1$ and $\mu_2= \mu_1\circ t^{-1}$.
Using the polar factorization theorem in the Euclidean case described above, we can provide an elegant new proof to the counterpart of Theorem \ref{thm:discrete} in the continuous case.
\begin{theorem}[Continuous case]
\label{thm:continuous}
    The map $t\left(\boldsymbol{x}\right)=\boldsymbol{T}\boldsymbol{x}+\boldsymbol{b}$ (where $\boldsymbol{T} > 0$) is the solution to an OT problem between a measure $\mu_1$ and a measure $\mu_2= \mu_1\circ t^{-1}$ with the cost $c\left(\boldsymbol{x},\boldsymbol{z}\right)=\left\Vert \boldsymbol{x}-\boldsymbol{z}\right\Vert _{2}^{2}$.
\end{theorem}
\begin{proof}
    Consider the function $\psi\left(\boldsymbol{x}\right)=\frac{1}{2}\boldsymbol{x}^{T}\boldsymbol{T}\boldsymbol{x}+\boldsymbol{b}^{T}\boldsymbol{x}$. Since $\boldsymbol{T}>0$, $\psi$ is $c$-convex with respect to the cost $c(\boldsymbol{x},\boldsymbol{z})=\|\boldsymbol{x}-\boldsymbol{z}\|^2_2$. In addition, $\nabla \psi = t$,  and therefore, by the polar factorization theorem, the map $t\left(\boldsymbol{x}\right)=\boldsymbol{T}\boldsymbol{x}+\boldsymbol{b}$ is the solution to the OT problem.
\end{proof}

{\color{black}
\subsection{Domain Adaptation in $\mathcal{P}_d$}
    \label{sub:DAonSPD}
}
In this work we focus on the cone manifold of SPD matrices
$\mathcal{P}_{d}$.
As a first example, consider the following linear map on $\mathcal{P}_d$, with a fixed, real-valued, invertible matrix $\boldsymbol{S}$:
\begin{equation}
    s\left(\boldsymbol{P}\right)=\boldsymbol{S}\boldsymbol{P}\boldsymbol{S}^{T}.    
    \label{eq:linear_map}
\end{equation}
Using the polar decomposition $\boldsymbol{S}=\boldsymbol{T}\boldsymbol{U}$, we can write $s\left(\boldsymbol{P}\right)=\boldsymbol{T}\boldsymbol{U}\boldsymbol{P}\boldsymbol{U}^{T}\boldsymbol{T}$.
We now show an analogous proposition to Theorem \ref{thm:continuous} for the the SPD case (with Euclidean cost function).
\begin{corollary}[SPD case with square Euclidean cost]
\label{cor:SPD}
    The linear map $\boldsymbol{Q} =  t\left(\boldsymbol{P}\right)=\boldsymbol{T}\boldsymbol{P}\boldsymbol{T}$ (where $\boldsymbol{T}>0$) is the solution to an OT problem between $\mu_1$ and $\mu_2= \mu_1\circ t^{-1}$ on $\mathcal{P}_d$ with {\color{black}the cost}
$c\left(\boldsymbol{P},\boldsymbol{Q}\right)=\left\Vert \boldsymbol{P}-\boldsymbol{Q}\right\Vert _{F}^{2}$  where $\boldsymbol{P},\boldsymbol{Q}\in\mathcal{P}_{d}$.
\end{corollary}
\begin{proof}
    Using the $\text{vec}\left(\cdot\right)$ operator we can write $\boldsymbol{q} = \left(\boldsymbol{T}\otimes\boldsymbol{T}\right)\boldsymbol{p}$, where $\boldsymbol{q}=\text{vec}\left(\boldsymbol{Q}\right)$, $\boldsymbol{p}=\text{vec}\left(\boldsymbol{P}\right)$, and $\otimes$ is the Kronecker product. Since $\boldsymbol{T} > 0$, then $\left(\boldsymbol{T}\otimes \boldsymbol{T}\right) > 0$ as well. Therefore, by Theorem \ref{thm:continuous}, $t\left(\boldsymbol{p}\right)=\left(\boldsymbol{T}\otimes\boldsymbol{T}\right)\boldsymbol{p}$ is the solution to the OT problem between $\mu_1$ and $\mu_2=\mu_1\circ t^{-1}$ with the given cost function.
\end{proof}

We conjecture that in this linear case the map $t$ is the solution to the OT problem on $\mathcal{P}_d$ with $c\left(\boldsymbol{P},\boldsymbol{Q}\right)=d_{R}^{2}\left(\boldsymbol{P},\boldsymbol{Q}\right)$ as well, i.e., with the cost defined through the Riemannian distance.
We leave the proof for future work, yet, in the following, we present empirical evidence supporting this conjecture.
More importantly, in the general {\color{black}case of non-linear maps}, the OT plan obtained using the Euclidean distance could be substantially different from the OT plan obtained using the Riemannian distance $d_R$. We postulate that the Riemannian distance $d_R$ is the one that should be used, as it demonstrates superior results in Section \ref{sec:ExperimentalResults}. 

    \begin{figure*}[t]
    \centering
        \includegraphics[width=.87\linewidth]{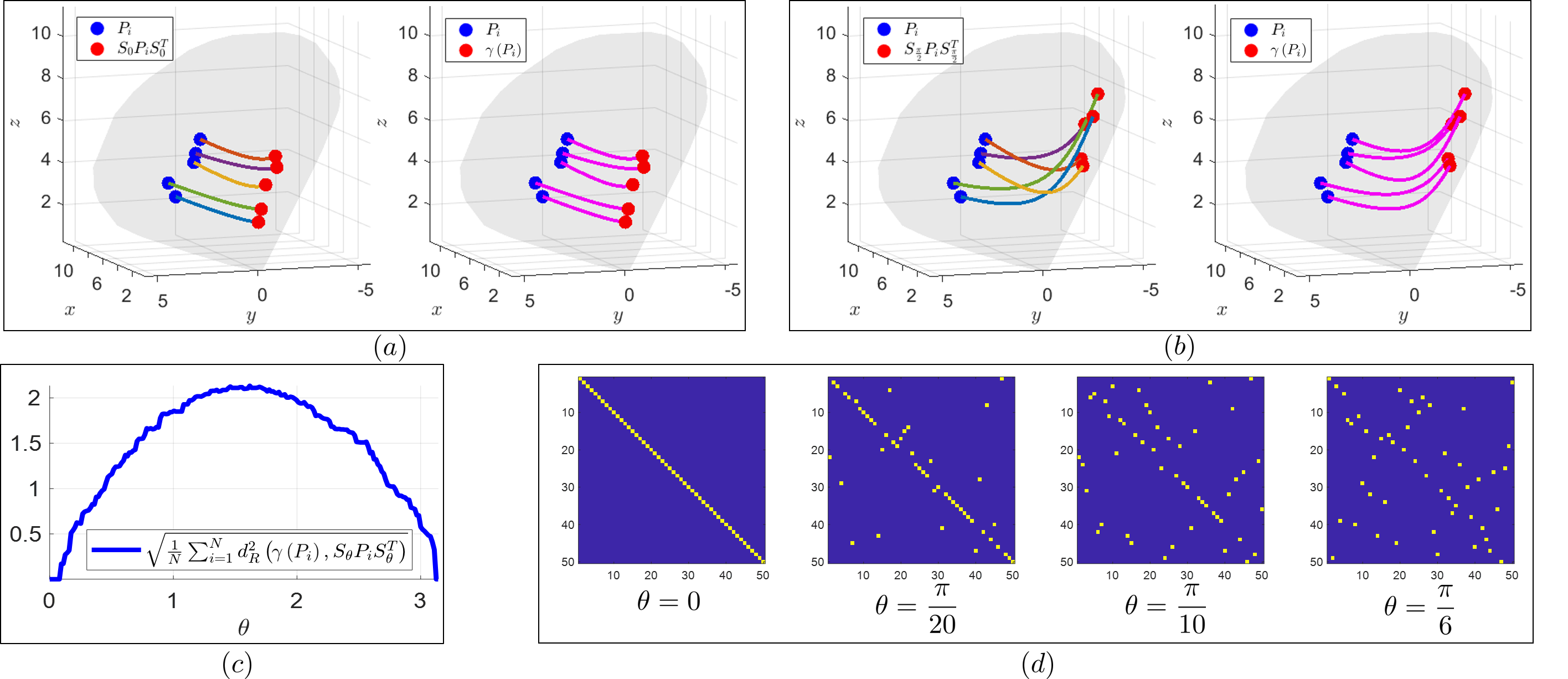}
        \caption{Results on the toy problem, please see \ref{sub:ToyProblemA} for details.}
        \label{fig:Toy}
    \end{figure*}
    
        \begin{figure}[t]
        \centering
        \includegraphics[width=1\columnwidth]{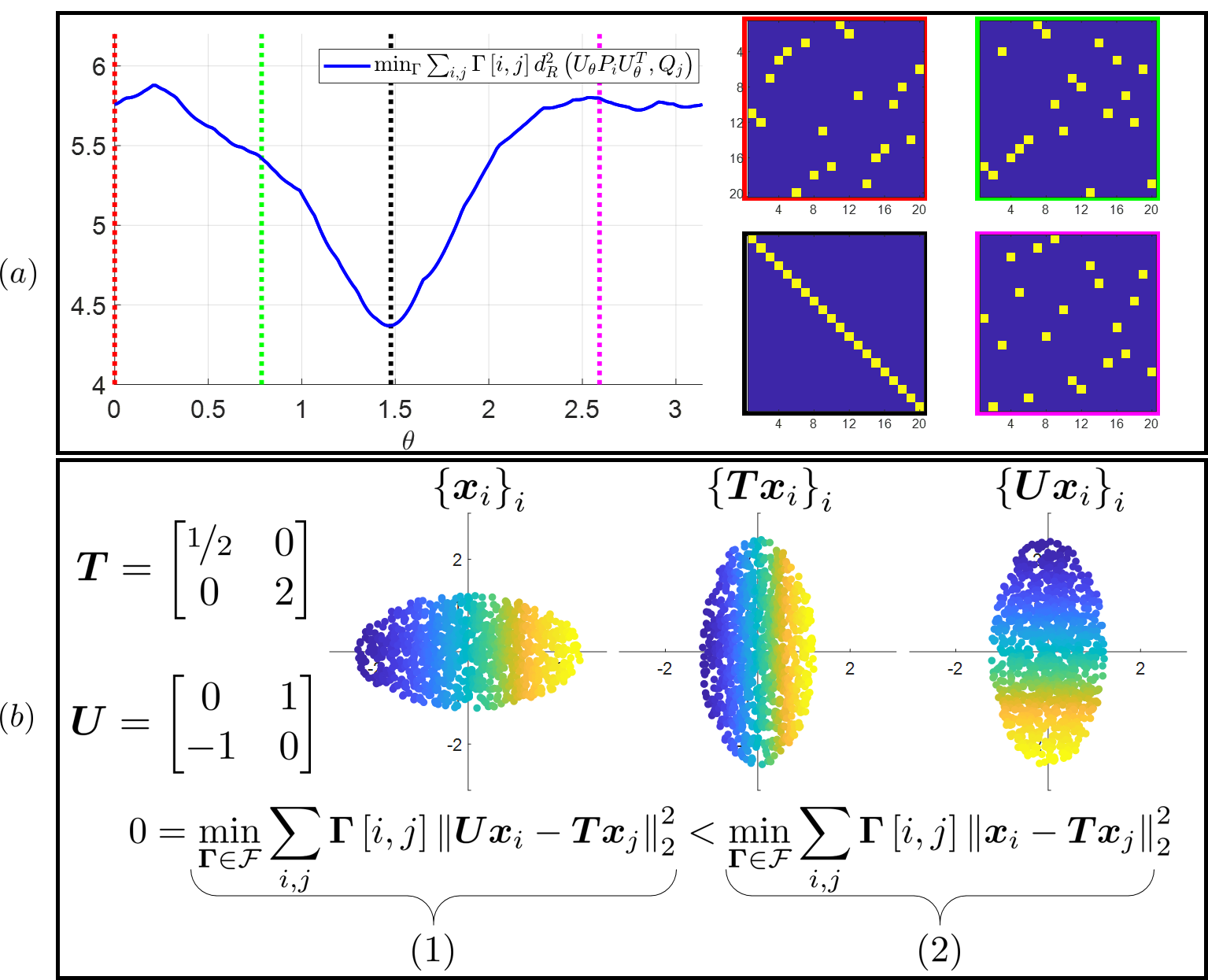}
        \caption
        {
            (a) Left -- the cost function value as a function of ${\theta}$.
            Right -- the transport plan $\boldsymbol{\Gamma}$ for four values of $\theta$. The red frame ($\theta = 0$) is equivalent to directly applying OT; whereas in the black frame ($\theta = \nicefrac{\pi}{2}$) we first multiply the matrices $\boldsymbol{P}_{i}$ with $\boldsymbol{U}_{\frac{\pi}{2}}$ and then solve the OT problem.
            (b) Consider the linear map $\boldsymbol{S}=\boldsymbol{T}$ specified above. Since $\boldsymbol{T}>0$, the theory guarantees that solving (2), will recover the true $\boldsymbol{S}$. Solving (1) with $\boldsymbol{U}$ specified above will not recover the true S. However the value of (1) is strictly smaller than (2). A similar example can be constructed in the SPD manifold as well.
        }
        \label{fig:ToyB}
    \end{figure}

\section{Experimental Results}
    \label{sec:ExperimentalResults} 
In this section we first present two toy examples, illustrating the matrix polar decomposition and the subsequent conjecture described in Section~\ref{sub:DAonSPD}. The second toy problem indicates a possible future direction when additional knowledge on the data is available. Then, we present results on two BCI data sets, where our method achieves state of the art performance. Our source code is available in\footnote{\href{https://github.com/oryair/SpdOptimalTransportDomainAdaptation}{SpdOptimalTransportDomainAdaptation GitHub}}.

\subsection{Toy Problems}
The two toy problems involve symmetric $2 \times 2$ matrices, because any symmetric $2 \times 2$ matrix $\boldsymbol{A}=\big[\begin{smallmatrix}x & y\\ y & z \end{smallmatrix}\big]$ can be visualized in $\mathbb{R}^3$ by plotting the elements $\left(x,y,z\right)$. In addition, $\boldsymbol{A}$ is SPD if and only if $x,z>0$ and $y^2<x z$. These conditions imply that the $2\times 2$ SPD matrices reside within a cone embedded in $\mathbb{R}^3$. In both toy problems, since the problems are computationally light and the purpose of this section is to illustrate the theoretical claims regarding OT, we apply Algorithm \ref{alg:OT} with uniform mass distribution and we solve the OT problem directly \eqref{eq:discrete ot}, without the Sinkhorn regularization.
    
    \label{sub:ToyProblemA}
Consider a $2\times 2$ SPD matrix $\boldsymbol{P}$ and a parametric map $s_\theta$ given by $s_{\theta}\left(\boldsymbol{P}\right)=\boldsymbol{S}_{\theta}\boldsymbol{P}\boldsymbol{S}_{\theta}^{T}$, where $\boldsymbol{S}=\boldsymbol{T}\boldsymbol{U}_{\theta}$, $\boldsymbol{T}=\big[\begin{smallmatrix}0.5 & -\nicefrac{1}{4}\\ -\nicefrac{1}{4} & 1 \end{smallmatrix}\big]>0$ and $\boldsymbol{U}_{\theta}=\big[\begin{smallmatrix}\cos(\theta)& \sin(\theta)\\ -\sin(\theta) & \cos(\theta) \end{smallmatrix}\big]$ is orthogonal.
First, we generate a source set of $5$ SPD matrices $\left\{ \boldsymbol{P}_{i}\right\} _{i=1}^{5}$ and apply $s_0$ and $s_{\nicefrac{\pi}{2}}$ to this set, namely $\left\{ s_{\theta}\left(\boldsymbol{P}_{i}\right)\right\} _{i=1}^{5}$ is the target set. We apply Algorithm \ref{alg:OT} to $\left\{ \boldsymbol{P}_{i}\right\} _{i=1}^{5}$ and $\{ s_\theta(\boldsymbol{P}_i)\}$ and obtain $\left\{ \gamma_{\theta}\left(\boldsymbol{P}_{i}\right)\right\} _{i=1}^{5}$.
The pairs $(\boldsymbol{P}_i,s_\theta(\boldsymbol{P}_i))$ embedded in $\mathbb{R}^3$ are depicted in Figure \ref{fig:Toy}(a) and (b). 
In the left plots, the pairs are connected by the geodesic curves. In the right plots, the curves represent the matching obtained by $\gamma_\theta$ computed using Algorithm \ref{alg:OT}.
In Figure \ref{fig:Toy}(a), where $\boldsymbol{S}_0 = \boldsymbol{T}>0$ and does not contain a orthogonal component, we observe that the obtained OT $\gamma_0$ indeed recovers the map $s_0$. Conversely, in Figure \ref{fig:Toy}(b), where $\boldsymbol{S}_{\nicefrac{\pi}{2}}$ has a non-trivial orthogonal component, we observe that the OT $\gamma_{\nicefrac{\pi}{2}}$ \textit{does not recover} the true matching.
    
To quantify this observation, we generate a set of $N = 50$ SPD matrices $\left\{ \boldsymbol{P}_{i}\right\} _{i=1}^{N}$ and consider maps $s_{\theta}$, where $\theta\in\left[0,\pi\right]$. For each value of $\theta$, we compute $\gamma_{\theta}$ using Algorithm \ref{alg:OT}. Figure \ref{fig:Toy}(c) displays the error between $s_{\theta}\left(\boldsymbol{P}_{i}\right)$ and the corresponding $\gamma_{\theta}\left(\boldsymbol{P}_{i}\right)$ defined by $\sqrt{\frac{1}{N}\sum_{i}^{N}d_{R}^{2}\left(\gamma_{\theta}\left(\boldsymbol{P}_{i}\right),s_{\theta}\left(\boldsymbol{P}_{i}\right)\right)}$. In addition, Figure \ref{fig:Toy}(d) presents the transport plans $\boldsymbol{\Gamma}$ obtained for $4$ values of $\theta$.
We observe that the ability of OT to recover $s_\theta$ depends on $\theta$; as the orthogonal component of $s_\theta$ is `less prominent' ($\theta$ is closer to $0$), the recovery becomes more accurate.

    \begin{figure*}
        \centering
        \includegraphics[width=1\linewidth]{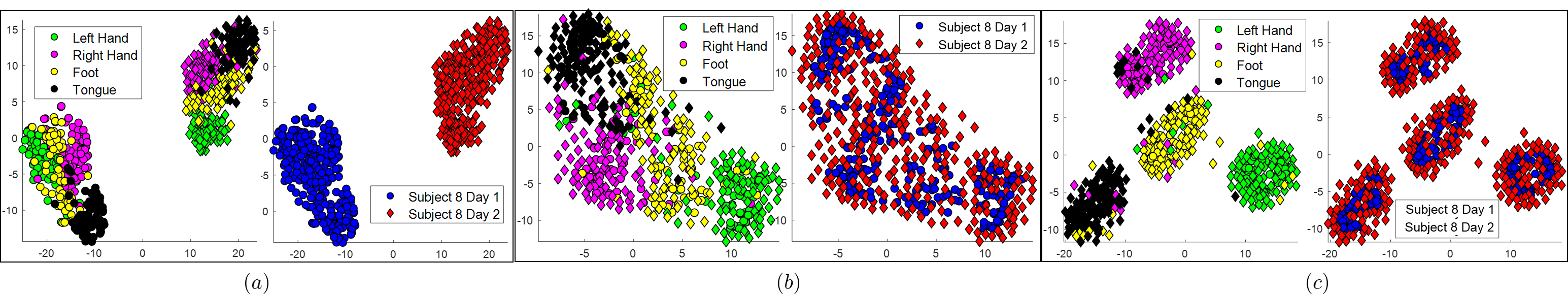}
        \caption{Cross-session adaptation in the motor imagery task. The t-SNE representation of all the SPD matrices of Subject $8$: (a) before Algorithm \ref{alg:OT}, (b) after Algorithm \ref{alg:OT} (label-free) and (c) after Algorithm \ref{alg:OT} (with source labels). Please see \ref{sub:BCI} for details.}
        \label{fig:BciSub8}
    \end{figure*}

    \label{sub:ToyProblemB}
    {\color{black}
For some linear maps $s$, solving the following non-convex optimization problem:
\begin{align}
    \label{eq:minU}
\underset{\boldsymbol{U}\in O}{\min}~\underset{\boldsymbol{\Gamma}\in \mathcal{F}}{\min}\sum_{i,j}\boldsymbol{\Gamma}\left[i,j\right]d_{R}^{2}(\boldsymbol{U}\boldsymbol{P}_{i}\boldsymbol{U}^{T},\boldsymbol{Q}_{j})
\end{align}
where $\boldsymbol{U}$ is orthogonal, may recover the true map $s$. This is demonstrated on a toy problem similar to the one described above. Consider a source set $\left\{ \boldsymbol{P}_{i}\right\} _{i=1}^{N=20}$ of SPD matrices in $\mathbb{R}^{2\times2}$ and a target set $\left\{ \boldsymbol{Q}_{i}\right\} _{i=1}^{N=20}$, s.t. $\boldsymbol{Q}_{i}=\boldsymbol{S}\boldsymbol{P}_{i}\boldsymbol{S}^{T}$ for some invertible matrix $\boldsymbol{S}$. 
Since in general $S$ is not an SPD matrix, applying OT to the two sets fails to recover the correct matching.
We heuristically solve \eqref{eq:minU} by: (i) sampling uniformly $\theta$ from $\left[0,2\pi\right]$, (ii) generating the corresponding rotation matrices $\boldsymbol{U}_{\theta}$, and (iii) solving the inner OT problem \eqref{eq:minU}, for each $\boldsymbol{U}_{\theta}$ candidate.
Note that in general, one should (exhaustively) search over a discretization of the entire space of orthogonal matrices.
Figure \ref{fig:ToyB}(a,left) presents the cost function value as a function of ${\theta}$. Figure \ref{fig:ToyB}(a,right) presents the transport plan $\boldsymbol{\Gamma}$ for four values of $\theta$.
In this example, directly applying OT to the two sets is equivalent to the result obtained for $\theta = 0$ (red frame) which does not recover the true map, whereas the min value of the (label-free) cost function \eqref{eq:minU} coincides with the true map (black frame). 

Seemingly, the above experiment contradicts our main claim about the fundamental limitation of OT for DA in a label-free setting, since the problem \eqref{eq:minU} is shown to coincide with the true map despite having a non-trivial orthogonal part.
However, even in this linear case, the fundamental limitation still holds and the minimum of the cost function is not guaranteed to coincide with the correct $u$. Figure \ref{fig:ToyB}(b) provides a simple counterexample in the linear Euclidean case.
Nevertheless, we believe this practice has potential when the setting is not strictly unsupervised and few source-target pairs are given. This will be further explored in future work.

    }

\subsection{Motor Imagery Task}
    \label{sub:BCI}
We use data from the BCI competition IV, see \cite{naeem2006seperability}, which have been previously addressed using the Riemannian geometry of SPD matrices by \cite{barachant2012multiclass,zanini2018transfer,yair2019parallel}. The dataset contains EEG recordings from 22 electrodes from 9 subjects, where each subject was recorded in 2 different days (sessions). In repeated trials, the subjects were asked to imagine performing one out of four possible movements: (i) right hand, (ii) left hand, (iii) both feet, and (iv) tongue. Overall, in a single day, each movement was repeated 72 times by each subject. The sampling rate is $250$Hz and each trial is 3 seconds long.
Here, we consider only 5 subjects out of the available 9 as in \cite{zanini2018transfer,yair2019parallel}. This is because the single-session single-subject classification results on data from each of the remaining 4 subjects were poor, see \cite{ang2012filter,barachant2012multiclass,zanini2018transfer}. Thus, DA is not relevant.
Let $\boldsymbol{X}_{i}^{\left(k,s\right)}\in\mathbb{R}^{22\times750}$ denote the data from the $i$th trial of the $k$th subject at the $s$th session, and let $y_{i}^{\left(k,s\right)}\in\left\{ \text{right hand, left hand, both feet, tongue}\right\}$ be the associated label. 
As in \cite{barachant2012multiclass, zanini2018transfer, yair2019parallel}, we preprocessed the data by applying a BPF with cutoff frequencies 8Hz and 30Hz.
For each trial we compute the empirical covariance $\boldsymbol{P}_{i}^{\left(k,s\right)} \in \mathbb{R}^{22 \times 22}$.

To illustrate the need for DA, Figure \ref{fig:BciSub8}(a) displays the 2-D t-SNE representation \cite{maaten2008visualizing} of all the SPD matrices of Subject $8$, namely $\left\{ \boldsymbol{P}_{i}^{\left(8,s\right)}\right\}, i=1,\ldots,288, s=1,2$, computed based on the pairwise {Riemannian} distances $d_R$\footnotemark. 
\footnotetext{The pairwise distances $d_R$ were used for the t-SNE representations of the covariance matrices in all other figures as well.}
We observe that the data is primarily clustered by session, whereas the clustering by the imagined movement is only secondary. This poses a challenge for classifiers trained on one session and applied to the other.
We adapt the SPD matrices of the first session by applying Algorithm \ref{alg:OT} twice. Once, without using any labels, and once, using the source labels as described in \ref{sub:ProposedMethod}. Let $\left\{ \widetilde{\boldsymbol{P}}_{i}^{\left(8,1\right)}\right\}$ denote the output of the algorithm. Figure \ref{fig:BciSub8}(b) depicts the 2-D t-SNE representation of $\left\{ \widetilde{\boldsymbol{P}}_{i}^{\left(8,1\right)}\right\}$ in the label-free setting, and similarly, Figure \ref{fig:BciSub8}(c) depicts the 2-D representation in the setting with the source labels. We observe that the data is primarily clustered by the imagined movement, whereas the session has only a mild effect. In addition, we observe the contribution of the known labels. For cross-subject classification we repeat the same process, but with two different subjects instead of the same subject and two different sessions. See Appendix C.2 for figures about multiple subjects adaptation.
    \begin{table*}
        \centering
        \caption{Classification accuracy in the motor imagery task. (l) stands for the version with source labels, (u) stands for the label-free (unlabeled version), and (E) stands for the Euclidean metric.}
        \label{tbl:BCI}
        \includegraphics[width=1\linewidth]{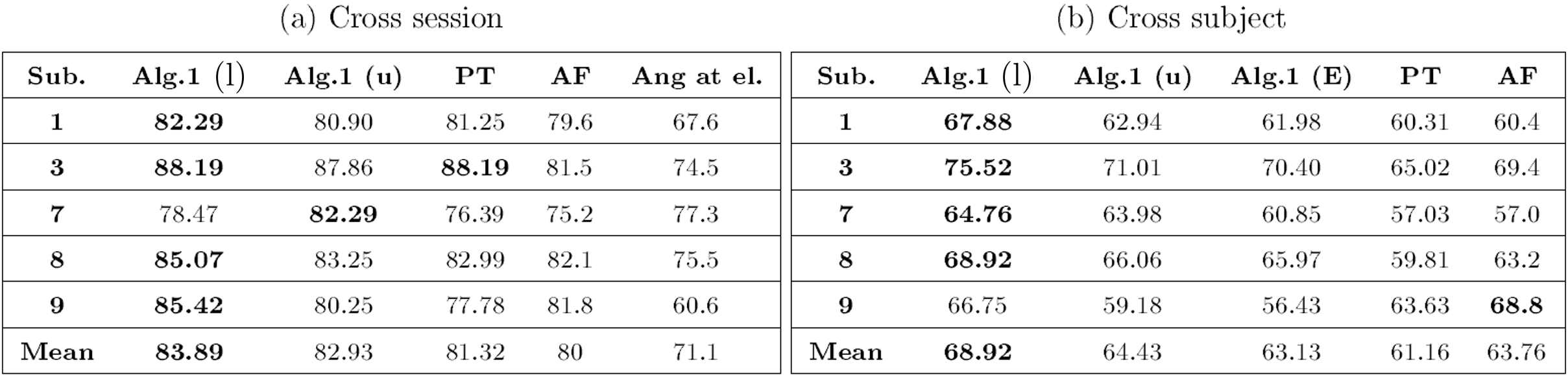}
    \end{table*}

To quantify the adaptation, we present the classification accuracy for cross-session and cross-subject classification.
For the cross-subject classification, we present the average result of each subject treated as a test set for all possible cross-subject pairs (see Appendix C.2 for the full detailed table).
We compare the performance to three methods: (i) the method in \cite{ang2012filter} which achieved the 1st place in the original competition, (ii) the Affine Transform (AT) proposed in \cite{zanini2018transfer}, and (iii) the Parallel Transport (PT) proposed in \cite{yair2019parallel}. For evaluation, we use a linear SVM classifier equipped with the Euclidean approximation in the tangest space as described in Appendix A.
Table \ref{tbl:BCI}(a) depicts the cross-session performance and Table \ref{tbl:BCI}(b) depicts the cross-subject performance. Note that the results in \cite{ang2012filter} are available only for the cross-session case. In addition, in the cross-subject case, we added a variant of Algorithm \ref{alg:OT}, which relies on the Euclidean distance instead of the Riemannian distance (labeled ``Euclid''). We observe that overall Algorithm \ref{alg:OT}, with the Riemannian metric, provides the best results. 

    \begin{figure*}
        \centering
        \includegraphics[width=1\linewidth]{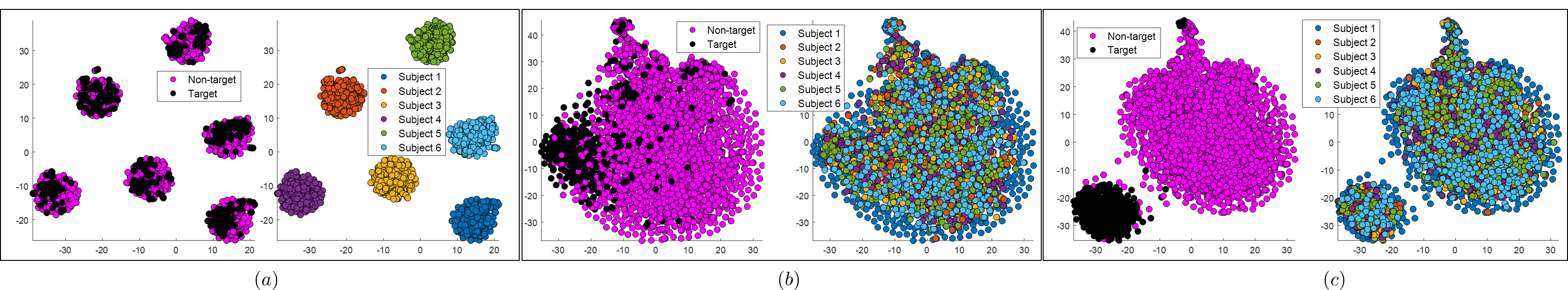}
        \caption{Cross-subject adaptation in the ERP P300 task. The t-SNE representation of the SPD matrices of three subjects: (a) before (b) after Algorithm \ref{alg:OT} (label-free) and (c) after Algorithm \ref{alg:OT} (with source labels). Please see Section \ref{sub:ERP} for details.}
        \label{fig:ERP}
    \end{figure*}

\subsection{P300 Event Related Potential Task}
    \label{sub:ERP}
We use data from the Brain Invaders experiment from GIPSA-lab. Here we only include a brief description; for more details see \cite{congedo2011brain}. In this experiment, subjects watched a screen with 36 objects flashing alternatively. Their task was to mentally count the number of flashes of the specific (a-priori known) target objects. Spotting a target generates a $P300$ wave, which is an Event Related Potential (ERP). Each subject watched $480$ trials, of which $80$ contained the target and the remaining $400$ did not. The data consist of EEG recordings from $16$ electrodes sampled at $512$Hz. The duration of each trial is one second. 

Note that the empirical covariance matrix is invariant to the temporal order of the samples, yet the ERP is a short local wave. Consequently, the covariance matrix does not capture sufficient information on the ERP in a given trial
Thus, instead of the standard covariance matrix, we use an augmented covariance matrix as proposed in \cite{barachantplug}. For more details on the pre-processing and the augmented covariance please see Appendix C.3.
We note that, in comparison to \cite{barachantplug} which used the test labels to computed the augmented covariances, we apply a completely label-free method. We denote the augmented covariance matrix associated with the $i$th trail of the $k$ subject by ${\boldsymbol{P}}_{i}^{\left(k\right)}\mathbb{R}^{32\times32}$ (after augmenting by 16 channels the original measured 16).

To illustrate the existing batch effect in the data, and subsequently, the need for DA, Figure \ref{fig:ERP}(a) displays the 2-D t-SNE representation of the covariance matrices from all six subjects. As in the previous application, we observe that the SPD matrices are primarily clustered by the subject.
Figure \ref{fig:ERP}(b) displays the covariance matrices from all six subjects after applying Algorithm \ref{alg:OT} in the label-free setting to adapt the SPD matrices of all subjects to the domain of the SPD matrices of Subject 1.
Similarly, Figure \ref{fig:ERP}(c) displays the data from all six subjects after applying Algorithm \ref{alg:OT} with sources labels to adapt the SPD matrices of all subjects to the domain of the SPD matrices of Subject 1, that is, using all labels except the labels of Subject 1 (the target domain). We observe that after the adaptation the SPD matrices are clustered by label of the trial and not by subject. In addition, similarly to the previous experiment, we clearly observe the contribution of the labels.


To quantify the adaptation, we computed the cross-subject classification precision, defined by $\text{pr}=\frac{\text{TP}}{\text{TP}+\text{FP}}$, where TP is the number of trials with target objects correctly classified, and FP is the number of trials without a target wrongly classified as with a target. Appendix C.3 contains detailed tables of the algorithm precision performance.
We compare Algorithm \ref{alg:OT} to the algorithms presented in \cite{zanini2018transfer} and in \cite{yair2019parallel}. For the comparison, we use the same Minimum Distance to Mean (MDM) classifier as in \cite{zanini2018transfer}, which was proposed in \cite{barachant2012multiclass}. Table \ref{tbl:ErpTotal} contains a summary of the classification precision obtained by Algorithm \ref{alg:OT} and the competing algorithms. The N\textbackslash{}A results were not reported in \cite{zanini2018transfer}, since these subjects were classified as ``bad'' subjects.
We observe that Algorithm \ref{alg:OT} provides the best results overall, and that the knowledge of the labels improves the classification.

    \begin{table}
        \centering
        \caption{Classification precision in the ERP P300 task. (l) stands for the version with the source labels, (u) stands for the label-free (unlabeled) version.}
        \label{tbl:ErpTotal}
        \includegraphics[width=.95\columnwidth]{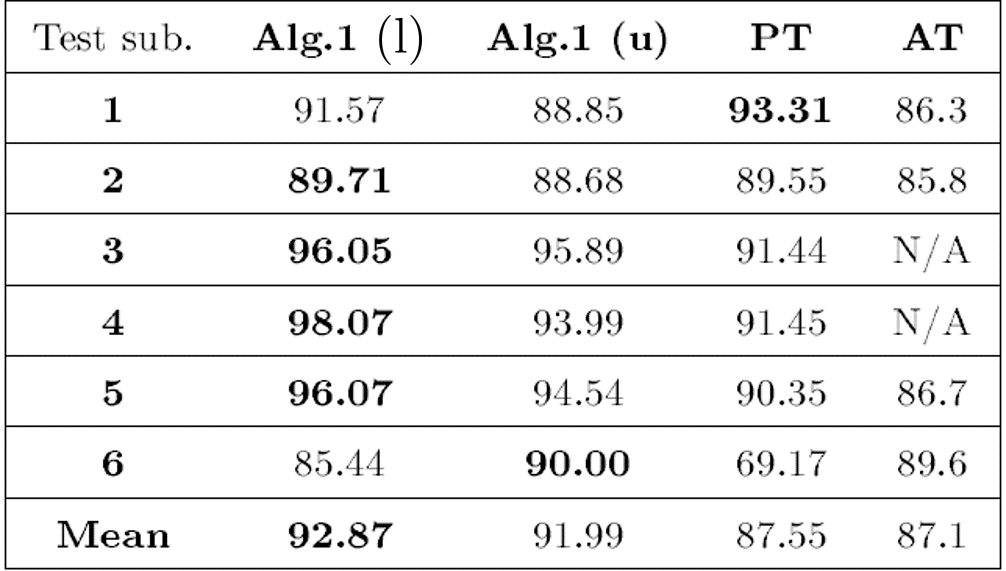}
    \end{table}
    

\section{Conclusion}
    \label{sec:Conclusion} 
    In this paper we considered DA on the manifold of SPD matrices using OT. Based on the polar factorization theorem, we presented a new analysis, highlighting the advantages and limitations of OT for DA. Our analysis applies to the general Riemannian setting. Particularly, on the SPD manifold, we showed that the transport map derived by OT is well-defined, and we proposed an algorithm for its efficient implementation. We tested our algorithm on two BCI tasks and demonstrated state of the art performance. 
    
    In the pure unsupervised setting, without any labels, the polar decomposition specifies the limitation of OT for DA. 
    We believe that the polar decomposition can be further exploited. Particularly, future work will address the recovery of the volume-preserving part, complementing the OT solution, given the true maps of a small number of points from the source set.

\bibliographystyle{IEEEtran}
\bibliography{Refs.bib}

\begin{thebibliography}{10}
\providecommand{\url}[1]{#1}
\csname url@samestyle\endcsname
\providecommand{\newblock}{\relax}
\providecommand{\bibinfo}[2]{#2}
\providecommand{\BIBentrySTDinterwordspacing}{\spaceskip=0pt\relax}
\providecommand{\BIBentryALTinterwordstretchfactor}{4}
\providecommand{\BIBentryALTinterwordspacing}{\spaceskip=\fontdimen2\font plus
\BIBentryALTinterwordstretchfactor\fontdimen3\font minus
  \fontdimen4\font\relax}
\providecommand{\BIBforeignlanguage}[2]{{%
\expandafter\ifx\csname l@#1\endcsname\relax
\typeout{** WARNING: IEEEtran.bst: No hyphenation pattern has been}%
\typeout{** loaded for the language `#1'. Using the pattern for}%
\typeout{** the default language instead.}%
\else
\language=\csname l@#1\endcsname
\fi
#2}}
\providecommand{\BIBdecl}{\relax}
\BIBdecl

\bibitem{daume2006domain}
H.~Daume~III and D.~Marcu, ``Domain adaptation for statistical classifiers,''
  \emph{Journal of artificial Intelligence research}, vol.~26, pp. 101--126,
  2006.

\bibitem{gong2012geodesic}
B.~Gong, Y.~Shi, F.~Sha, and K.~Grauman, ``Geodesic flow kernel for
  unsupervised domain adaptation,'' in \emph{2012 IEEE Conference on Computer
  Vision and Pattern Recognition}.\hskip 1em plus 0.5em minus 0.4em\relax IEEE,
  2012, pp. 2066--2073.

\bibitem{pan2009survey}
S.~J. Pan and Q.~Yang, ``A survey on transfer learning,'' \emph{IEEE
  Transactions on knowledge and data engineering}, vol.~22, no.~10, pp.
  1345--1359, 2009.

\bibitem{pennec2006riemannian}
X.~Pennec, P.~Fillard, and N.~Ayache, ``A riemannian framework for tensor
  computing,'' \emph{International Journal of computer vision}, vol.~66, no.~1,
  pp. 41--66, 2006.

\bibitem{arsigny2007geometric}
V.~Arsigny, P.~Fillard, X.~Pennec, and N.~Ayache, ``Geometric means in a novel
  vector space structure on symmetric positive-definite matrices,'' \emph{SIAM
  journal on matrix analysis and applications}, vol.~29, no.~1, pp. 328--347,
  2007.

\bibitem{wang2012covariance}
R.~Wang, H.~Guo, L.~S. Davis, and Q.~Dai, ``Covariance discriminative learning:
  A natural and efficient approach to image set classification,'' in \emph{2012
  IEEE Conference on Computer Vision and Pattern Recognition}.\hskip 1em plus
  0.5em minus 0.4em\relax IEEE, 2012, pp. 2496--2503.

\bibitem{jayasumana2013kernel}
S.~Jayasumana, R.~Hartley, M.~Salzmann, H.~Li, and M.~Harandi, ``Kernel methods
  on the riemannian manifold of symmetric positive definite matrices,'' in
  \emph{Proceedings of the IEEE Conference on Computer Vision and Pattern
  Recognition}, 2013, pp. 73--80.

\bibitem{barachant2013classification}
A.~Barachant, S.~Bonnet, M.~Congedo, and C.~Jutten, ``Classification of
  covariance matrices using a {R}iemannian-based kernel for bci applications,''
  \emph{Neurocomputing}, vol. 112, pp. 172--178, 2013.

\bibitem{wang2015beyond}
L.~Wang, J.~Zhang, L.~Zhou, C.~Tang, and W.~Li, ``Beyond covariance: Feature
  representation with nonlinear kernel matrices,'' in \emph{Proceedings of the
  IEEE International Conference on Computer Vision}, 2015, pp. 4570--4578.

\bibitem{freifeld2014model}
O.~Freifeld, S.~Hauberg, and M.~J. Black, ``Model transport: Towards scalable
  transfer learning on manifolds,'' in \emph{Proceedings of the IEEE Conference
  on Computer Vision and Pattern Recognition}, 2014, pp. 1378--1385.

\bibitem{courty2017optimal}
N.~Courty, R.~Flamary, D.~Tuia, and A.~Rakotomamonjy, ``Optimal transport for
  domain adaptation,'' \emph{IEEE transactions on pattern analysis and machine
  intelligence}, vol.~39, no.~9, pp. 1853--1865, 2017.

\bibitem{courty2014domain}
N.~Courty, R.~Flamary, and D.~Tuia, ``Domain adaptation with regularized
  optimal transport,'' in \emph{Joint European Conference on Machine Learning
  and Knowledge Discovery in Databases}.\hskip 1em plus 0.5em minus 0.4em\relax
  Springer, 2014, pp. 274--289.

\bibitem{bhatia2009positive}
R.~Bhatia, \emph{Positive definite matrices}.\hskip 1em plus 0.5em minus
  0.4em\relax Princeton university press, 2009, vol.~16.

\bibitem{villani-2009}
C.~Villani, \emph{Optimal Transport}.\hskip 1em plus 0.5em minus 0.4em\relax
  Springer Berlin Heidelberg, 2009.

\bibitem{fathi-2010}
A.~Fathi and A.~Figalli, ``Optimal transportation on non-compact manifolds,''
  \emph{Israel Journal of Mathematics}, vol. 175, no.~1, pp. 1--59, jan 2010.

\bibitem{cuturi2013sinkhorn}
M.~Cuturi, ``Sinkhorn distances: Lightspeed computation of optimal transport,''
  in \emph{Advances in neural information processing systems}, 2013, pp.
  2292--2300.

\bibitem{mccann2001polar}
R.~McCann, ``Polar factorization of maps on riemannian manifolds,''
  \emph{Geometric \& Functional Analysis GAFA}, vol.~11, no.~3, pp. 589--608,
  2001.

\bibitem{brenier-1991}
Y.~Brenier, ``Polar factorization and monotone rearrangement of vector-valued
  functions,'' \emph{Communications on Pure and Applied Mathematics}, vol.~44,
  no.~4, pp. 375--417, jun 1991.

\bibitem{modin-2017}
K.~Modin, ``Geometry of matrix decompositions seen through optimal transport
  and information geometry,'' \emph{Journal of Geometric Mechanics}, vol.~9,
  no.~3, pp. 335--390, jun 2017.

\bibitem{brenier1987decomposition}
Y.~Brenier, ``D{\'e}composition polaire et r{\'e}arrangement monotone des
  champs de vecteurs,'' \emph{CR Acad. Sci. Paris S{\'e}r. I Math.}, vol. 305,
  pp. 805--808, 1987.

\bibitem{naeem2006seperability}
M.~Naeem, C.~Brunner, R.~Leeb, B.~Graimann, and G.~Pfurtscheller,
  ``Seperability of four-class motor imagery data using independent components
  analysis,'' \emph{Journal of neural engineering}, vol.~3, no.~3, p. 208,
  2006.

\bibitem{barachant2012multiclass}
A.~Barachant, S.~Bonnet, M.~Congedo, and C.~Jutten, ``Multiclass
  brain--computer interface classification by riemannian geometry,'' \emph{IEEE
  Transactions on Biomedical Engineering}, vol.~59, no.~4, pp. 920--928, 2012.

\bibitem{zanini2018transfer}
P.~Zanini, M.~Congedo, C.~Jutten, S.~Said, and Y.~Berthoumieu, ``Transfer
  learning: a riemannian geometry framework with applications to
  brain--computer interfaces,'' \emph{IEEE Transactions on Biomedical
  Engineering}, vol.~65, no.~5, pp. 1107--1116, 2018.

\bibitem{yair2019parallel}
O.~Yair, M.~Ben-Chen, and R.~Talmon, ``Parallel transport on the cone manifold
  of spd matrices for domain adaptation,'' \emph{IEEE Transactions on Signal
  Processing}, vol.~67, no.~7, pp. 1797--1811, 2019.

\bibitem{ang2012filter}
K.~Ang, Z.~Y. Chin, C.~Wang, C.~Guan, and H.~Zhang, ``Filter bank common
  spatial pattern algorithm on bci competition iv datasets 2a and 2b,''
  \emph{Frontiers in neuroscience}, vol.~6, p.~39, 2012.

\bibitem{maaten2008visualizing}
L.~v.~d. Maaten and G.~Hinton, ``Visualizing data using t-sne,'' \emph{Journal
  of machine learning research}, vol.~9, no. Nov, pp. 2579--2605, 2008.

\bibitem{congedo2011brain}
M.~Congedo, M.~Goyat, N.~Tarrin, G.~Ionescu, L.~Varnet, B.~Rivet, R.~Phlypo,
  N.~Jrad, M.~Acquadro, and C.~Jutten, ````brain invaders'': a prototype of an
  open-source p300-based video game working with the openvibe platform,'' in
  \emph{5th International Brain-Computer Interface Conference 2011 (BCI 2011)},
  2011, pp. 280--283.

\bibitem{barachantplug}
A.~Barachant and M.~Congedo, ``A plug\&play {P}300 {BCI} using information
  geometry,'' \emph{arXiv preprint arXiv:1409.0107}, 2014.

\end{thebibliography}

\end{document}


\setcounter{equation}{13}

\title{Domain Adaptation with Optimal Transport on the Manifold of SPD matrices -- Appendix}

\author{Or~Yair,
        Felix Dietrich,
        Ioannis G. Kevrekidis,
        and~Ronen Talmon
\IEEEcompsocitemizethanks{\IEEEcompsocthanksitem O. Yair and R. Talmon are with the Viterbi Faculty of Electrical Engineering, Technion, Israel Institute of Technology.\protect\\
E-mail: oryair@campus.technion.ac.il
\IEEEcompsocthanksitem Felix Dietrich is with the Technical University of Munich.
\IEEEcompsocthanksitem Ioannis G. Kevrekidis is with the Department of Chemical and Biomolecular Engineering and of Applied Mathematics and Statistics, Johns Hopkins University.}
\thanks{The work of O. Yair and R. Talmon was supported by the European Union's Horizon 2020 research grant agreement 802735. The work of F. Dietrich and I. G. Kevrekidis was partially supported by DARPA and by the Army Research Office MURI.}
}





\maketitle

\IEEEdisplaynontitleabstractindextext
\IEEEpeerreviewmaketitle


\appendices
\section{On the Cone Manifold of SPD Matrices -- Additional Information}
    \label{app:SPD}

In this section, we provide additional information about the SPD manifold.
The Exponential map from $\mathcal{T}_{\boldsymbol{P}}\mathcal{P}_{d}$ to $\mathcal{P}_d$ is given explicitly by
    \begin{align}
    \label{eq:Exp}
    \boldsymbol{P}_{i} =\text{Exp}_{\boldsymbol{P}}(\boldsymbol{A}_{i})=\boldsymbol{P}^{\frac{1}{2}}\exp\big(\boldsymbol{P}^{-\frac{1}{2}}\boldsymbol{A}_{i}\boldsymbol{P}^{-\frac{1}{2}}\big)\boldsymbol{P}^{\frac{1}{2}}.
    \end{align}
    where $\exp\left(\boldsymbol{P}\right)$ is the matrix exponential.
    The Logarithm map from $\mathcal{P}_d$ to $\mathcal{T}_{\boldsymbol{P}}\mathcal{P}_{d}$ is given explicitly by 
    \begin{align}
    \label{eq:Log}
    \boldsymbol{A}_{i}  =\text{Log}_{\boldsymbol{P}}(\boldsymbol{P}_{i})=\boldsymbol{P}^{\frac{1}{2}}\log\big(\boldsymbol{P}^{-\frac{1}{2}}\boldsymbol{P}_{i}\boldsymbol{P}^{-\frac{1}{2}}\big)\boldsymbol{P}^{\frac{1}{2}}.
    \end{align}
    
    The geodesic metric can be efficiently computed by
    $$d_{R}^{2}\left(\boldsymbol{P},\boldsymbol{Q}\right)=\sum_{i=1}^{d}\log^{2}\left(\lambda_{i}\left(\boldsymbol{Q}^{-1}\boldsymbol{P}\right)\right)$$
    where $\lambda_{i}\left(\boldsymbol{Q}^{-1}\boldsymbol{P}\right)$
    is the $i$th eigenvalue of $\boldsymbol{Q}^{-1}\boldsymbol{P}$,
    or the $i$th general eigenvalue of the pair $\left(\boldsymbol{P,}\boldsymbol{Q}\right)$.

    The Riemannian mean $\overline{\boldsymbol{P}}$ of a set $\left\{ \boldsymbol{P}_{i} | \boldsymbol{P}_{i}\in\mathcal{P}_d\right\} $ is defined using the Fr\'echet mean:
    \begin{equation}
        \label{eq:riemannian_mean}
        \overline{\boldsymbol{P}}=\arg\min_{\boldsymbol{P}\in\mathcal{P}_d}\sideset{}{_{i}}\sum d_{R}^{2}\big(\boldsymbol{P},\boldsymbol{P}_{i}\big).
    \end{equation}

    The optimization problem \eqref{eq:riemannian_mean} is strictly convex and thus, it is well defined, see \cite{bhatia2009positive}, and can be solved by an iterative procedure. Algorithm \ref{alg:WeightedMean} computes the \textit{weighted} Riemannian mean. Given a set $\left\{ \boldsymbol{P}_{i}\in\mathcal{P}_d\right\} $ and its mean $\overline{\boldsymbol{P}}$, a commonly-used Euclidean approximation of the Riemannian distances on $\mathcal{P}_d$ in the neighborhood of $\overline{\boldsymbol{P}}$ is given by
    \begin{align}
        \label{eq:TangentApproxiamtion}
        d_{R}^{2}\big(\boldsymbol{P}_{i},\boldsymbol{P}_{j}\big)\approx\big\Vert     \widetilde{\boldsymbol{A}}_{i}-\widetilde{\boldsymbol{A}}_{j}\big\Vert _{F}^{2}\,\,,
    \end{align}
    where $\widetilde{\boldsymbol{A}}_{i}=\overline{\boldsymbol{P}}^{-\tfrac{1}{2}}\text{Log}_{\overline{\boldsymbol{P}}}(\boldsymbol{P}_{i})\overline{\boldsymbol{P}}^{-\tfrac{1}{2}}=\log\left(\overline{\boldsymbol{P}}^{-\frac{1}{2}}\boldsymbol{P}_{i}\overline{\boldsymbol{P}}^{-\frac{1}{2}}\right)$ and $\widetilde{\boldsymbol{A}}_j$ is defined analogously.
    For more details on the accuracy of this approximation, see \cite{tuzel2008pedestrian}.

\begin{algorithm}
    \renewcommand{\thealgorithm}{2}
   \caption{Weighted Riemannian mean}
   \label{alg:WeightedMean}
    
    \textbf{\uline{Input}}\textbf{:} a set of SPD matrices $\left\{ \boldsymbol{P}_{i}\right\} _{i=1}^{N}$ and non-negative weights $\left\{w_{i}\right\} _{i=1}^{N}$ such that $\sum_{i} w_{i}=1$.
        
    \textbf{\uline{Output}}\textbf{:} the weighted Riemannian mean $\overline{\boldsymbol{P}}$ satisfying $\overline{\boldsymbol{P}}=\arg\min_{\boldsymbol{P}\in\mathcal{P}_d}\sum_{i}w_{i}d_{R}^{2}\left(\boldsymbol{P},\boldsymbol{P}_{i}\right)$.
    
    \begin{algorithmic}[1]
        \STATE \textbf{set}: $\overline{\boldsymbol{P}} \gets \frac{1}{N}\sum_{i=1}^{N}w_{i}\boldsymbol{P}_{i}$. \hfill \COMMENT{starting point}
        
        \STATE \textbf{do}:
        
        \STATE \qquad \textbf{update}: $\overline{\boldsymbol{S}} \gets \frac{1}{N}\sum_{i=1}^{N}w_{i}\text{Log}_{\overline{\boldsymbol{P}}}\left(\boldsymbol{P}_{i}\right)$. \hfill
        \COMMENT{weighted Euclidean mean in $\mathcal{T}_{\overline{\boldsymbol{P}}}\mathcal{P}_{d}$, see \eqref{eq:Log}}
        
        \STATE \qquad \textbf{update}: $\overline{\boldsymbol{P}} \gets \text{Exp}_{\overline{\boldsymbol{P}}}\left(\overline{\boldsymbol{S}}\right)$.\hfill
        \COMMENT{see \eqref{eq:Exp}}
        
        \STATE \textbf{while} $\left\Vert \overline{\boldsymbol{S}}\right\Vert _{F}>\epsilon$ \hfill
        \COMMENT{$\left\Vert \cdot\right\Vert _{F}$ is the Frobenius norm}

    \end{algorithmic}
\end{algorithm}

\section{Regularized Optimal Transport Algorithms}
    \label{app:OtAlgorithms}
In Algorithm 1 in the paper, Step 4 solves the OT problem.
In this section, we describe two possible fast implementations of a regularized version of the classical OT.
The first implementation is completely label-free, whereas, the second implementation supports the case, where the labels of the \emph{source set} are known.

\subsection{Classical Optimal Transport}
Given two discrete density vectors, source $\boldsymbol{c} \in \mathbb{R}^{N_1}$ and target $\boldsymbol{r} \in \mathbb{R}^{N_2}$, as well as the cost matrix $\boldsymbol{C} \in \mathbb{R}^{N_1 \times N_2}$, the discrete version of optimal transport is the following optimization problem:
\begin{equation}
    \label{eq:OT}
   \arg \min_{\boldsymbol{\Gamma}}\left\langle \boldsymbol{\Gamma},\boldsymbol{C}\right\rangle 
\end{equation}
such that $\boldsymbol{\Gamma}\boldsymbol{1}=\boldsymbol{c}$ and $\boldsymbol{\Gamma}^T\boldsymbol{1}=\boldsymbol{r}$, where $\boldsymbol{1}$ is a vector of all ones in a suitable dimension.

\subsection{Sinkhorn Optimal Transport}
Often, the computation cost of \eqref{eq:OT} becomes extremely high when the dimensions $N_1$ and $N_2$ exceed several hundreds. \cite{cuturi2013sinkhorn} proposed to solve the optimal transport problem with a regularization term based on entropy, which can be computed with the Sinkhorn’s matrix scaling algorithm at a speed that is several orders of magnitude faster than that of classical optimal transport solvers. The proposed regularized problem is:
\begin{equation}
    \label{eq:Sinkhorn2}
    \arg \min_{\boldsymbol{\Gamma}}\left\langle \boldsymbol{\Gamma},\boldsymbol{C}\right\rangle -\frac{1}{\lambda}h\left(\boldsymbol{\Gamma}\right)    
\end{equation}
where $h\left(\boldsymbol{\Gamma}\right)=-\sum_{i=1}^{N_{1}}\sum_{j=1}^{N_{2}}\boldsymbol{\Gamma}\left[i,j\right]\log\left(\boldsymbol{\Gamma}\left[i,j\right]\right)$ is the entropy of $\boldsymbol{\Gamma}$. We note that in the toy problem in Section 4.1 we assign large values to $\lambda$, so that the regularized problem \eqref{eq:Sinkhorn2} is similar to the classical problem \eqref{eq:OT}. For the real data sets in Section 4.2 and Section 4.3, we set $\lambda$ adaptively by
$$ \lambda=\frac{1}{2m^{2}}$$
where $m=0.05\cdot\text{median}\left\{ \boldsymbol{C}\left[i,j\right]\right\} _{i,j}$.
We implemented Algorithm \ref{alg:Sinkhorn}, which was proposed by \cite{cuturi2013sinkhorn} and solves 
\eqref{eq:Sinkhorn2}, as outlined below.

\begin{algorithm}
    \renewcommand{\thealgorithm}{3}
    \caption{Sinkhorn optimal transport proposed by \cite{cuturi2013sinkhorn}}
    \label{alg:Sinkhorn}

    \textbf{\uline{Input}:} $\boldsymbol{C} \in \mathbb{R}^{N_1 \times N_2}$, $\lambda \in \mathbb{R}$, $\boldsymbol{c} \in \mathbb{R}^{N_1}$, and $\boldsymbol{r} \in \mathbb{R}^{N_2}$.
        
    \textbf{\uline{Output}}\textbf{:} the transport plan $\boldsymbol{\Gamma} \in \mathbb{R}^{N_1 \times N_2}$ (solution of \eqref{eq:Sinkhorn2}).
    
    \begin{algorithmic}[1]
        
        \STATE \textbf{set}: $\boldsymbol{K}\left[i,j\right] \gets \exp\left(-\lambda\boldsymbol{C}\left[i,j\right]\right)$
        \hfill \COMMENT{$\forall i \in\left\{ 1,\dots,N_{1}\right\}$ and $\forall j \in\left\{ 1,\dots,N_{2}\right\}$}
        
        \STATE \textbf{set}: $\boldsymbol{u}\left[i\right] \gets \frac{1}{N_{1}}$.
        \hfill \COMMENT{$\forall i\in\left\{ 1,\dots,N_{1}\right\}$}
        
        \STATE \textbf{set}: $\widetilde{\boldsymbol{K}}\left[i,j\right]\gets\frac{\boldsymbol{K}\left[i,j\right]}{\boldsymbol{c}\left[i\right]}$.
        \hfill \COMMENT{$\forall i \in\left\{ 1,\dots,N_{1}\right\}$ and $\forall j \in\left\{ 1,\dots,N_{2}\right\}$}
        
        \STATE \textbf{while} $u$ changes \textbf{do}: 
        
        \STATE \qquad $\boldsymbol{z}\left[j\right]\gets\frac{\boldsymbol{r}\left[j\right]}{\left(\boldsymbol{K}^{T}\boldsymbol{u}\right)\left[j\right]}$
        \hfill \COMMENT{$\forall j\in\left\{ 1,\dots,N_{2}\right\}$}
        
        
        \STATE \qquad \textbf{update}: $\boldsymbol{u}\left[i\right]=\frac{1}{\left(\widetilde{\boldsymbol{K}}\boldsymbol{z}\right)\left[i\right]}$
        \hfill \COMMENT{$\forall i\in\left\{ 1,\dots,N_{1}\right\}$}
        
        \STATE \textbf{end while}
        
        \STATE \textbf{set}: $\boldsymbol{v}\left[j\right] \gets \frac{\boldsymbol{r}\left[j\right]}{\left(\boldsymbol{K}^{T}\boldsymbol{u}\right)\left[j\right]}$
        \hfill \COMMENT{$\forall j\in\left\{ 1,\dots,N_{2}\right\}$}
        
        \STATE \textbf{set}: $\boldsymbol{\Gamma}\gets\text{diag}\left(\boldsymbol{u}\right)\boldsymbol{K}\text{diag}\left(\boldsymbol{v}\right)$

    \end{algorithmic}
\end{algorithm}    

\subsection{Sinkhorn Optimal Transport with Source Labels}

Consider now a setting consisting of two sets of SPD matrices in $\mathcal{P}_d$. The source set $\left\{ \boldsymbol{P}_{i},y_{i}\right\} _{i=1}^{N_{1}}$ is given with labels $y_{i}\in\mathcal{Y}$, where $\mathcal{Y}$ is the set of all possible labels, and the target set $\left\{ \boldsymbol{Q}_{i}\right\} _{i=1}^{N_{2}}$ is unlabeled (i.e., the labels are unknown).
We set $\boldsymbol{C}_{0}\left[i,j\right]=d_{R}^{2}\left(\boldsymbol{P}_{i},\boldsymbol{Q}_{j}\right)$. \cite{courty2014domain} presented a modification of the Sinkhorn optimal transport problem with an additional label regularization term and derived a new efficient algorithm to solve the following problem:
\begin{equation}
    \label{eq:RegOT2}
    \arg\min_{\boldsymbol{\Gamma}}\left\langle \boldsymbol{\Gamma},\boldsymbol{C}_0\right\rangle -\frac{1}{\lambda}h\left(\boldsymbol{\Gamma}\right)+\eta\sum_{j=1}^{N_{2}}\sum_{y=1}^{\left|\mathcal{Y}\right|}\left\Vert \boldsymbol{\Gamma}\left(\mathcal{I}_{y},j\right)\right\Vert _{q}^{p}
\end{equation}
where $\mathcal{I}_{y}$ is a set containing the indices of the source points associated with the label $y\in\mathcal{Y}$, $\boldsymbol{\Gamma}\left(\mathcal{I}_{y},j\right)$ is a vector consisting of entries from the $j$th column of $\boldsymbol{\Gamma}$ associated with the label $y$ (i.e., from rows $\mathcal{I}_{y}$) and $\left\Vert \cdot\right\Vert _{q}^{p}$ is the $L_q$ norm to the power of $p$. See \cite{courty2014domain} for more details.
We implemented Algorithm \ref{alg:RegOT}, which was proposed by \cite{courty2014domain} and solves
\eqref{eq:RegOT2} for $q=1$ and $p=2$.

\begin{algorithm}
    \renewcommand{\thealgorithm}{4}
    \caption{Sinkhorn optimal transport with labels proposed by \cite{courty2014domain}}
    \label{alg:RegOT}

    \textbf{\uline{Input}:} $\boldsymbol{C}_0 \in \mathbb{R}^{N_1 \times N_2}$, $\lambda \in \mathbb{R}$, $\boldsymbol{c} \in \mathbb{R}^{N_1}$, $\boldsymbol{r} \in \mathbb{R}^{N_2}$, and the source labels $\left\{ y_{i}\right\} _{i=1}^{N_{1}}$.
        
    \textbf{\uline{Output}}\textbf{:} the transport plan $\boldsymbol{\Gamma}$ (solution of \eqref{eq:RegOT2}).
    
    \begin{algorithmic}[1]
        
        \STATE \textbf{initialize}: $\boldsymbol{G}\gets\boldsymbol{0}$ 
        \hfill \COMMENT{$\boldsymbol{G}\in\mathbb{R}^{N_{1}\times N_{2}}$}
        
        \STATE \textbf{do}
        
        \STATE \qquad \textbf{update} $\boldsymbol{C}\gets\boldsymbol{C}_{0}+\boldsymbol{G}$
        
        \STATE \qquad \textbf{update} $\boldsymbol{\Gamma}\gets\text{SinkhornOptimalTransport}\left(\boldsymbol{C},\lambda,\boldsymbol{c},\boldsymbol{r}\right)$
        \hfill \COMMENT{Algorithm \ref{alg:Sinkhorn}}
        
        \STATE \qquad \textbf{update} $\boldsymbol{G}$ according to
        \hfill \COMMENT{$\forall y$ and $\forall j$}
        $$\boldsymbol{G}\left(\mathcal{I}_{y},j\right)\gets p\cdot\left(\left\Vert {\boldsymbol{\Gamma}}\left(\mathcal{I}_{y},j\right)\right\Vert +\epsilon\right)^{p-1}$$
            
        \STATE \textbf{while} $\boldsymbol{\Gamma}$ changes
 
     \end{algorithmic}
\end{algorithm}

\section{Additional Results}
    \label{app:AdditionalResults}

\subsection{Application: Illustration on Simulated Data}
    \label{app:DAonSPD}
In the following, we assume that the two sets of SPD matrices $\left\{ \boldsymbol{P}_{i}\in\mathcal{P}_{d}\right\} _{i=1}^{N}$ and $\left\{ \boldsymbol{Q}_{i}\in\mathcal{P}_{d}\right\} _{i=1}^{N}$ are covariance matrices originating from the raw data sets $\left\{ \boldsymbol{X}_{i}\in\mathbb{\mathbb{R}}^{d\times M}\right\} _{i=1}^{N}$ and $\left\{ \boldsymbol{Z}_{i}\in\mathbb{\mathbb{R}}^{d\times M}\right\} _{i=1}^{N}$. That is, $\boldsymbol{Z}_{i}=s\left(\boldsymbol{X}_{i}\right)$, $\boldsymbol{P}_{i}=\text{cov}\left(\boldsymbol{X}_{i}\right)=\frac{1}{M-1}\boldsymbol{X}_{i}\boldsymbol{X}_{i}^{T}$  (assuming zero mean), and $\boldsymbol{Q}_{i}=\text{cov}\left(\boldsymbol{Z}_{i}\right)=\text{cov}\left(s\left(\boldsymbol{X}_{i}\right)\right)$ where here $s$ is the mapping between the raw data in hte Euclidean space. Alternately, one can replace the $\text{cov}\left(\cdot\right)$ function with some other SPD kernel.

In general, under this setting, there are three natural alternatives for applying OT and obtaining the transport plan for domain adaptation. First, applying OT to the raw data with the Euclidean distance as the cost. Second, applying OT to the covariance matrices with the Euclidean distance. Third, applying OT to the covariance matrices with the Riemannian distance $d_R$.
We postulate that the first two methods ignore the geometry of the data sets leading to subpar performance. Conversely,  by considering OT with the Riemannian distance $d_R$ on $\mathcal{P}_d$, we incorporate the geometric structure.

To emphasize this claim, consider the following $N=40$ time series pairs $\left(\boldsymbol{x}^{\left(i\right)}\left(t\right),\boldsymbol{z}^{\left(i\right)}\left(t\right)\right)$:
$$\boldsymbol{x}^{\left(i\right)}\left(t\right)=\left[\begin{matrix}x_{1}^{\left(i\right)}\left(t\right)\\
\vdots\\
x_{5}^{\left(i\right)}\left(t\right)
\end{matrix}\right],\boldsymbol{z}^{\left(i\right)}\left(t\right)=\left[\begin{matrix}z_{1}^{\left(i\right)}\left(t\right)\\
\vdots\\
z_{5}^{\left(i\right)}\left(t\right)
\end{matrix}\right],i\in\left\{ 1,2,\dots,40\right\} $$
where, for each time stamp $t_{0}$, $\boldsymbol{x}^{\left(i\right)}\left(t_{0}\right)$,$\boldsymbol{z}^{\left(i\right)}\left(t_{0}\right)\in\mathbb{R}^{5}$ and they are given by
$$\begin{matrix}x_{j}^{\left(i\right)}\left(t\right)=a_{j}^{\left(i\right)}\cos\left(f_{j}^{\left(i\right)}t+\theta_{j}^{\left(i\right)}\right)+n_{j}^{\left(i\right)}\\
z_{j}^{\left(i\right)}\left(t\right)=a_{j}^{\left(i\right)}\cos\left(f_{j}^{\left(i\right)}t+\phi_{j}^{\left(i\right)}\right)+\eta_{j}^{\left(i\right)}
\end{matrix},\qquad j\in\left\{ 1,2,\dots,5\right\} , $$
Namely, $x_{j}^{\left(i\right)}$ and $z_{j}^{\left(i\right)}$ are cosine function with the same frequency $f_{j}^{\left(i\right)}$ and the same amplitude $a_{j}^{\left(i\right)}$.
The two signals are differ in phase and noise.
Specifically,
$a_{j}^{\left(i\right)},f_{j}^{\left(i\right)}\sim U\left[0,20\right]$,
$\theta_{j}^{\left(i\right)},\phi_{j}^{\left(i\right)}\sim U\left[0,2\pi\right]$, and $n_{j}^{\left(i\right)},\eta_{j}^{\left(i\right)}\sim\mathcal{N}\left(0,1\right)$.
We denote
\begin{align*}
\boldsymbol{X}_{i} & =\left[\begin{matrix}\boldsymbol{x}^{\left(i\right)}\left(0\right) & \boldsymbol{x}^{\left(i\right)}\left(T_{s}\right) & \boldsymbol{x}^{\left(i\right)}\left(2T_{s}\right) & \cdots & \boldsymbol{x}^{\left(i\right)}\left(1\right)\end{matrix}\right]\in\mathbb{R}^{5\times T}\\
\boldsymbol{Z}_{j} & =\left[\begin{matrix}\boldsymbol{z}^{\left(i\right)}\left(0\right) & \boldsymbol{z}^{\left(i\right)}\left(T_{s}\right) & \boldsymbol{z}^{\left(i\right)}\left(2T_{s}\right) & \cdots & \boldsymbol{z}^{\left(i\right)}\left(1\right)\end{matrix}\right]\in\mathbb{R}^{5\times T}
\end{align*}

where $T_{s}=0.01$ and $T=101$.
Figure \ref{fig:Comparison}(a) displays realizations of $\boldsymbol{X}_{1}$ and $\boldsymbol{Z}_{1}$.
Let $\boldsymbol{P}_{i}=\text{cov}\left(\boldsymbol{X}_{i}\right)$ and $\boldsymbol{Q}_{i}=\text{cov}\left(\boldsymbol{Z}_{i}\right)$.
Given the two sets $\left\{ \boldsymbol{X}_{i}\right\} _{i}$ and $\left\{ \boldsymbol{Z}_{i}\right\} _{i}$, we apply the three configurations of the OT to: (i) the two sets with Euclidean distance, (ii) to their covariance matrices with Euclidean distance and (iii) to their covariance matrices with the Riemannian distance $d_R$.
Figure \ref{fig:Comparison}(b-d) presents the obtained transport plan for each configuration. The first configuration (raw data), Fig. \ref{fig:Comparison}(b), is sensitive to time shifts and thus it provides a poor matching. The covariance matrices capture a more global structure, and thus, are less sensitive to time shifts. Hence, the second configuration (covariances with Euclidean distance) in Fig. \ref{fig:Comparison}(c) performs better than the first configuration, but not as good as the last configuration (covariances with Riemannian distance) in Fig. \ref{fig:Comparison}(d), which provides a near perfect matching.

Therefore, as demonstrate in the main paper (see Section 4), the Riemannian distance facilitates meaningful comparisons and enables us to achieve state of the art performance in applications.

\setcounter{figure}{5}
\begin{figure*}
    \centering
    \includegraphics[width=.6\linewidth]{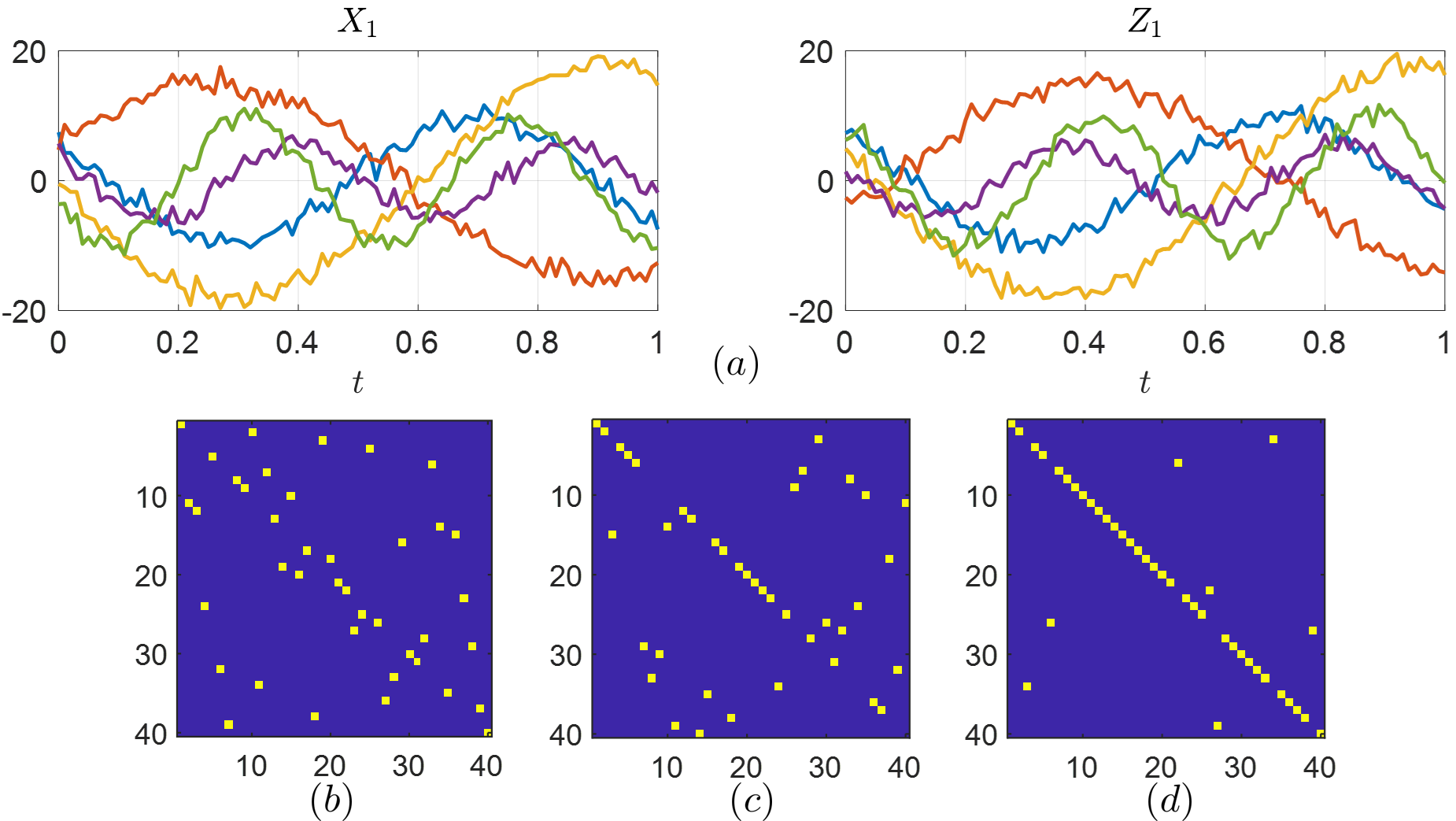}
    \caption{(a) Realizations of $\boldsymbol{X}_1$ and $\boldsymbol{Z}_1$.
             Bottom, the transport plans obtained by OT applied to:
             (b) The two sets $\left\{ \boldsymbol{X}_{i}\right\} _{i}$ and $\left\{ \boldsymbol{Z}_{i}\right\} _{i}$ with Euclidean distance.
             (c) The covariance matrices $\left\{ \boldsymbol{P}_{i}\right\} _{i}$ and $\left\{ \boldsymbol{Q}_{i}\right\} _{i}$ with Euclidean distance.
             (d) The covariance matrices $\left\{ \boldsymbol{P}_{i}\right\} _{i}$ and $\left\{ \boldsymbol{Q}_{i}\right\} _{i}$ with the Riemannian distance $d_R$.}
    \label{fig:Comparison}
\end{figure*}


\subsection{Motor imagery task}
    \label{app:AdditionalResultsBCI}

Prior to computing the covariance matrices, we applied a band pass filter with cutoff frequencies 8Hz and 30Hz.
This preprocessing was performed in the previous works \cite{barachant2012multiclass, zanini2018transfer, yair2019parallel}, where this data set was used.

\begin{figure*}
        \centering
        \includegraphics[width=1\linewidth]{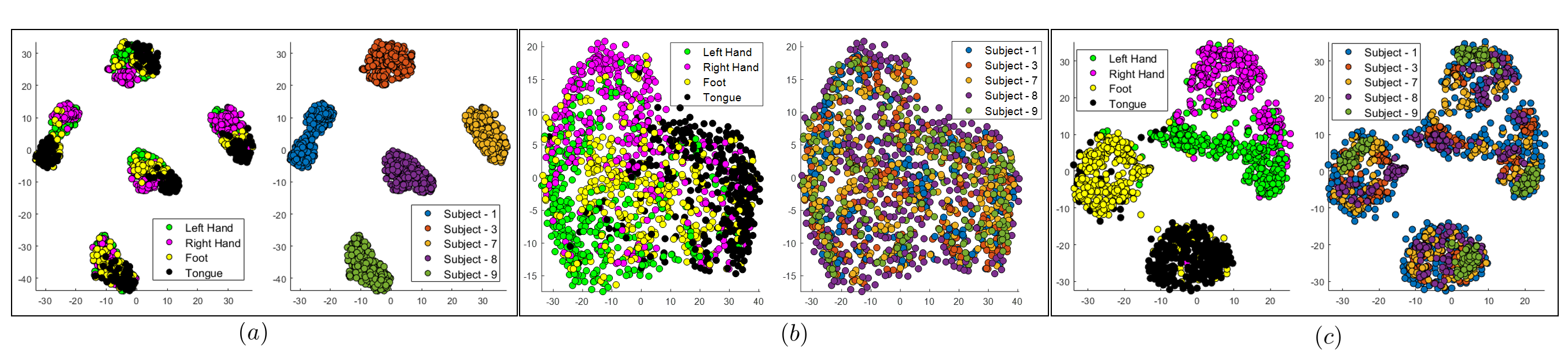}
        \caption{Multi-subject domain adaptation of the data from the BCI motor imagery task. (a) 2-dimensional t-SNE representation of all the SPD matrices of all five subjects (in Session 1). (b) 2-dimensional t-SNE representation after applying Algorithm 1 (label-free). (c) 2-dimensional t-SNE representation after applying Algorithm 1 (with source labels).}
        \label{fig:BciAll}
\end{figure*}

{\color{black} In Section 4}, we provide an illustrative result to the cross-session adaptation. We now test the adaptation of five sets corresponding to all subjects\footnote{As reported in the paper, we consider only 5 subject out of the available 9. See the explanation in Section 4.2 for details.}.
One set corresponding to Subject 1 was set as the reference set. Then, we applied Algorithm 1 to the remaining sets of the 4 subjects, mapping them one by one to the reference set.
Figure \ref{fig:BciAll}(a) displays the 2-dimensional t-SNE representation to the SPD matrices of all five subjects (in session 1). Figure \ref{fig:BciAll}(b) displays the 2-dimensional t-SNE representation of the SPD matrices after applying Algorithm 1 (label-free). Figure \ref{fig:BciAll}(c) displays the 2-dimensional t-SNE representation of the SPD matrices after applying Algorithm 1 (with source labels). We observe that after the adaptation the SPD matrices are clustered by label of the trial and not by subject. In addition, similarly to the experiments in the main paper, we clearly observe the contribution of the labels.

Finally, Table \ref{tbl:BCI2} provides the cross-subject classification accuracy, extending Table 1(b), which appears in Section 4.2. The left part of the table displays the pairwise classification accuracy when one subject used a train set and another a test set.
We observe that overall Algorithm 1, with the Riemannian metric, provides the best results
\setcounter{table}{2}
\begin{table*}
    \caption{Cross-subject classification accuracy in the motor imagery task. The four right-most columns consist of the mean results obtained per subject. Alg. 1 stands for Algorithm 1, Alg. 1 (Euclid) stands for Algorithm 1 when the Euclidean distance is used instead of the Riemannian distance $d_R$. PT stands for domain adaptation using Parallel Transport by \cite{yair2019parallel}. AT stands for the domain adaptation using Affine Transform by \cite{zanini2018transfer}. The $\pm 7$ is the overall standard deviation of Algorithm 1.}
    \label{tbl:BCI2}
    \begin{center}
    \begin{tabular}{cccccccccc}
    \textbf{\backslashbox{\scriptsize{}{Test sub.}}{\scriptsize{}{Train sub.}}} & \textbf{1} & \textbf{3} & \textbf{7} & \textbf{8} & \textbf{9} & \textbf{Alg. 1} & \textbf{Alg. 1 (Euclid)} & \textbf{PT} & \textbf{AT}\tabularnewline
    \hline 
    \textbf{1} &  & 78.13 & 67.36 & 70.14 & 55.90 & \textbf{67.88} & 61.98 & 60.31 & 60.4\tabularnewline
    \hline 
    \textbf{3} & 78.13 &  & 79.17 & 78.13 & 66.67 & \textbf{75.52} & 70.40 & 65.02 & 69.4\tabularnewline
    \hline 
    \textbf{7} & 63.54 & 70.49 &  & 62.50 & 68.40 & \textbf{64.76} & 60.85 & 57.03 & 57.0\tabularnewline
    \hline 
    \textbf{8} & 68.06 & 72.22 & 67.01 &  & 66.32 & \textbf{68.92} & 65.97 & 59.81 & 63.2\tabularnewline
    \hline 
    \textbf{9} & 52.43 & 67.36 & 72.22 & 75.00 &  & 66.75 & 56.43 & 63.63 & \textbf{68.8}\tabularnewline
    \hline 
    \textbf{Mean} &  &  &  &  &  & $\textbf{68.92}\pm 7$ & 63.13 & 61.16 & 63.76\tabularnewline
    \end{tabular}
    \par\end{center}
\end{table*}

\begin{table*}
    \centering
    \caption{Classification precision of Algorithm 1 (label-free) in the ERP P300 task}
    \label{tbl:ErpSinkhorn}
    \begin{tabular}{cccccccc}
\textbf{\backslashbox{\scriptsize{}{Test sub.}}{\scriptsize{}{Train sub.}}} & \textbf{1} & \textbf{2} & \textbf{3} & \textbf{4} & \textbf{5} & \textbf{6} & \textbf{Mean}\tabularnewline
\hline 
\textbf{1} &  & 88.06 & 91.30 & 88 & 90.62 & 86.30 & \textbf{88.85}\tabularnewline
\hline 
\textbf{2} & 90.90 &  & 92.53 & 88.57 & 86.11 & 85.29 & \textbf{88.68}\tabularnewline
\hline 
\textbf{3} & 97.10 & 97.22 &  & 97.33 & 93.15 & 94.66 & \textbf{95.89}\tabularnewline
\hline 
\textbf{4} & 94.91 & 95.08 & 95.08 &  & 88.23 & 96.66 & \textbf{93.99}\tabularnewline
\hline 
\textbf{5} & 97.22 & 97.22 & 86.30 & 97.29 &  & 94.66 & \textbf{94.54}\tabularnewline
\hline 
\textbf{6} & 94.91 & 94.66 & 91.30 & 92.53 & 76.60 &  & \textbf{90.00}\tabularnewline
\hline 
\textbf{Mean} & \textbf{95.01} & \textbf{94.45} & \textbf{91.30} & \textbf{92.74} & \textbf{86.94} & \textbf{91.51} & \textbf{91.99$\pm$}3.9\tabularnewline
\end{tabular}
\end{table*}

\begin{table*}
    \centering
    \caption{Classification precision of Algorithm 1 (with source labels) in the ERP P300 task}
    \label{tbl:ErpReg}
    \begin{tabular}{cccccccc}
\textbf{\backslashbox{\scriptsize{}{Test sub.}}{\scriptsize{}{Train sub.}}} & \textbf{1} & \textbf{2} & \textbf{3} & \textbf{4} & \textbf{5} & \textbf{6} & \textbf{Mean}\tabularnewline
\hline 
\textbf{1} &  & 93.94 & 93.94 & 90.91 & 92.68 & 86.36 & \textbf{91.57}\tabularnewline
\hline 
\textbf{2} & 96.77 &  & 90.91 & 91.43 & 93.94 & 75.51 & \textbf{89.71}\tabularnewline
\hline 
\textbf{3} & 97.73 & 97.87 &  & 95.56 & 95.56 & 95.35 & \textbf{96.05}\tabularnewline
\hline 
\textbf{4} & 97.06 & 100 & 100 &  & 95.83 & 97.44 & \textbf{98.07}\tabularnewline
\hline 
\textbf{5} & 98.21 & 96.49 & 93.94 & 98.36 &  & 93.33 & \textbf{96.07}\tabularnewline
\hline 
\textbf{6} & 88.37 & 85.71 & 76.60 & 84.00 & 92.50 &  & \textbf{85.44}\tabularnewline
\hline 
\textbf{Mean} & \textbf{95.62} & \textbf{94.80} & \textbf{91.07} & \textbf{92.05} & \textbf{94.10} & \textbf{89.59} & \textbf{92.87$\pm$}6.1\tabularnewline
\end{tabular}
\end{table*}

\subsection{Event related potential P300 task}
    \label{app:AdditionalResultsERP}
Prior to computing the covariance matrices, we applied a band pass filter with cutoff frequencies 1Hz and 24Hz.
This preprocessing is implemented in the py.BI.EEG.2013-GIPSA github code. A similar preprocessing was also applied in \cite{barachantplug}.
We denote the data from $i$th trial of the $k$th subject by $\boldsymbol{X}_{i}^{\left(k\right)}\in\mathbb{R}^{16\times512}$.
Note that the empirical covariance matrix is invariant to the temporal order of the samples, yet the ERP is a short local wave. Consequently, the covariance matrix does not capture sufficient information on the ERP in a given trial. Thus, instead of the standard covariance matrix, we use an augmented covariance matrix as proposed by \cite{barachantplug} by concatenating artificial channels to the data. While \cite{zanini2018transfer} used the true hidden labels for that purpose, we apply a practical approach which does not require access to any hidden labels.
In \cite{zanini2018transfer}, the augmented covariance matrices were computed as follows. 
Let ${\mathcal{T}^{(k)}}$ be the trials of the $k$th subject containing an ERP response. 
The knowledge of $\mathcal{T}^{(k)}$ is unavailable since this set contains the labels we try to classify.
If ${\mathcal{T}^{(k)}}$ were given, the average ERP response can be computed by:
\[
\overline{\boldsymbol{X}}^{\left(k\right)}=\frac{1}{\left|{\mathcal{T}^{(k)}}\right|}\sum_{i\in{\mathcal{T}^{(k)}}}\boldsymbol{X}_{i}^{\left(k\right)}\in\mathbb{R}^{16\times512}.
\]
Next, the average response $\overline{\boldsymbol{X}}^{\left(k\right)}$ is appended to the data of each trial, obtaining: 
\[\left[\begin{matrix}\overline{\boldsymbol{X}}^{\left(k\right)}\\
\boldsymbol{X}_{i}^{\left(k\right)}
\end{matrix}\right]\in\mathbb{R}^{32\times512}.
\]
Then, the empirical covariance ${\boldsymbol{P}}_{i}^{\left(k\right)}$ of the augmented data, now of dimension ${32\times32}$, is computed.

Here, we propose a different approach, which does not require the knowledge of ${\mathcal{T}^{(k)}}$.
For each subject $k$, we apply the ERP detector proposed by \cite{hoffmann2005boosting} to all 480 trials. The overall accuracy of this ERP detector was very low and it cannot be used as a stand-alone method for classification of data from multiple subjects. 
However, when considering only the $15$ trials for each subject $k$ with the highest confidence score provided by the detector as trials with an ERP response, we approximate a subset of ${\mathcal{T}^{(k)}}$. In turn, this subset is used to compute the augmented covariance matrices similarly to the procedure described above. We remark that on average $12$ out of the $15$ trials indeed include an ERP response.

It is important to note that the entire data set contains 24 subjects. We test our algorithm on 6 six subjects discussed in \cite{zanini2018transfer} (4 ``good'' subjects and 2 ``bad'' ones), and we use the remaining subjects to train the P300 detector. Consequently, the P300 detector and the proposed domain adaptation algorithm do not share the same data.

Finally, Table \ref{tbl:ErpSinkhorn} provides the cross-subject classification accuracy, extending the label-free column in Table 2, which appears in Section 4.3. Similarly, Table \ref{tbl:ErpReg} extends the results with the source labels, which appear in Table 2 as well.
We observe that Algorithm 1 provides the best results overall, and that the knowledge of the labels improves the classification.

\bibliographystyle{IEEEtran}
\bibliography{Refs.bib}